\documentclass{article}

\usepackage[final]{ARLET_2025}

\usepackage[utf8]{inputenc} 
\usepackage[T1]{fontenc}    
\usepackage{titletoc}
\usepackage{tikz}
\usetikzlibrary{positioning, shapes.geometric}

\usepackage{hyperref}       
\usepackage{url}            
\usepackage{booktabs}       
\usepackage{amsfonts}       
\usepackage{nicefrac}       
\usepackage{microtype}      
\usepackage{xcolor}         

\usepackage{amsmath,amsfonts,bm}

\usepackage{amssymb}           
\usepackage{enumitem}          
\usepackage{amsthm}            
\usepackage{xcolor}            
\usepackage{framed}            
\usepackage{tcolorbox}         
\usepackage{mathtools}         
\usepackage{thm-restate}       
\usepackage{authblk}           

\def\eqref#1{equation~\ref{#1}}

\def\1{\bm{1}}

\DeclareMathAlphabet{\mathsfit}{\encodingdefault}{\sfdefault}{m}{sl}
\SetMathAlphabet{\mathsfit}{bold}{\encodingdefault}{\sfdefault}{bx}{n}

\newcommand\Regret{\mathrm{Regret}}

\newtheorem{theorem}{Theorem}
\newtheorem{lemma}[theorem]{Lemma}  

\newtheorem{proposition}[theorem]{Proposition}

\newtheorem{definition}[theorem]{Definition}

\DeclareMathOperator*{\argmax}{arg\,max}
\DeclareMathOperator*{\argmin}{arg\,min}

\usepackage{graphicx}
\usepackage{float}
\usepackage{caption}
\usepackage{subcaption}

\usepackage[ruled,vlined]{algorithm2e}
\usepackage{listings}

\definecolor{codegreen}{rgb}{0,0.6,0}           
\definecolor{codegray}{rgb}{0.5,0.5,0.5}       
\definecolor{codepurple}{rgb}{0.58,0,0.82}     
\definecolor{codeblue}{rgb}{0,0,0.8}           
\definecolor{backcolour}{rgb}{0.95,0.97,1}     

\lstdefinestyle{mystyle}{
    backgroundcolor=\color{backcolour},   
    commentstyle=\color{codegreen},
    keywordstyle=\color{magenta},
    numberstyle=\tiny\color{codegray},
    stringstyle=\color{codepurple},
    basicstyle=\ttfamily\footnotesize\color{black},
    breakatwhitespace=false,         
    breaklines=true,                 
    captionpos=b,                    
    keepspaces=true,                 
    numbers=left,                    
    numbersep=5pt,                  
    showspaces=false,                
    showstringspaces=false,
    showtabs=false,                  
    tabsize=2
}

\tikzset{
    conv/.style={rectangle, draw, fill=red!30, minimum width=2cm, minimum height=0.8cm, text centered},
    conv2/.style={rectangle, draw, fill=red!30, minimum width=4cm, minimum height=0.8cm, text centered},
    conv3/.style={rectangle, draw, fill=red!30, minimum width=6cm, minimum height=0.8cm, text centered},
    relu/.style={rectangle, draw, fill=blue!30, minimum width=1.5cm, minimum height=0.6cm, text centered},
    fc/.style={rectangle, draw, fill=green!50, minimum width=2cm, minimum height=0.8cm, text centered},
    fc2/.style={rectangle, draw, fill=green!50, minimum width=4cm, minimum height=0.8cm, text centered},
    fc3/.style={rectangle, draw, fill=green!50, minimum width=5cm, minimum height=0.8cm, text centered},
    flatten/.style={rectangle, draw, fill=yellow!50, minimum width=2cm, minimum height=0.8cm, text centered},
    title/.style={font=\bfseries, align=center}
}

\title{Improved Training Mechanism for \\ Reinforcement Learning via Online Model Selection}

\author{%
  \textbf{Aida Afshar$^1$, Aldo Pacchiano$^{1,2}$} \\
  $^1$Boston University\\
  $^2$Broad Institute of MIT and Harvard\\
  \texttt{aafshar@bu.edu} \\
}

\begin{document}

\maketitle

\begin{abstract}
We study the problem of online model selection in reinforcement learning, where the selector has access to a class of reinforcement learning agents and learns to adaptively select the agent with the right configuration. Our goal is to establish the improved efficiency and performance gains achieved by integrating online model selection methods into reinforcement learning training procedures.  We examine the theoretical characterizations that are effective for identifying the right configuration in practice, and address three practical criteria from a theoretical perspective: 1) Efficient resource allocation, 2) Adaptation under non-stationary dynamics, and 3) Training stability across different seeds. Our theoretical results are accompanied by empirical evidence from various model selection tasks in reinforcement learning, including neural architecture selection, step-size selection, and self model selection.
\end{abstract}

\section{Introduction}
A major effort in theoretical analysis of reinforcement learning algorithms is towards providing guarantees on the regret or sample complexity \citep{agarwal2019reinforcement, foster2023foundations}. These guarantees often rely on assumptions and problem-specific constant that are left to be configured at deployment. Therefore, one would only expect the guarantee to hold in practice under the assumption that the agent is well-specified; the configuration of the algorithm is consistent with the true nature of the problem instance. Otherwise, in case of what is known as misspecification \citep{foster2020adapting}, we observe a mismatch between the theoretical guarantee and the realized performance of the agent. 

Model Selection offers a remedy to the theory-practice mismatch that arises from misspecification. Given access to a set of base agents, the model selection algorithm, which we will refer to as the selector, interacts with an environment and learns to select the right model for the problem at hand. The algorithmic goal of model selection is to guarantee that the meta-learning algorithm for the selector has performance comparable to the best \textit{solo} base agent, without knowing a priori which agent is best suited for the problem at hand.

While theoretical model selection guarantees have been established across various sequential decision-making problems \citep{pacchiano2021model, lee2022oracle, liu2025model, cutkosky2021dynamic}, this work focuses specifically on online Reinforcement Learning. We propose a training mechanism for reinforcement learning that serves as a more efficient alternative to standard hyperparameter selection procedures. We analyze the performance of this mechanism across three model selection tasks and establish corresponding theoretical properties for data-driven model selection:

\begin{enumerate}
    \item We derive the relationship between the realized performance of each base agent and the compute that is allocated to them by the selector. We show that for model selection algorithms that satisfy a balancing property, the selector learns to direct more compute to base agents with better realized performance.
    \item Under non-stationary dynamics where the optimal base agent is likely to change, data-driven model selection can adapt to the new optimal choice. 
    \item In scenarios where the performance of the RL algorithm is sensitive to the randomness of the initialization, online model selection can stabilize the training by leveraging multiple initializations of the algorithm within a single run. We prove a simple argument showing how integrating model selection strategies into RL training procedure reduces the likelihood of failure.
\end{enumerate}

We leverage model selection methods that neither require structural assumptions about the RL agents nor their theoretical regret bounds. These algorithms use the \emph{realized rewards} of agents to do model selection in a data-driven manner \cite{dann2024data}. Thus, our training mechanism is algorithm-agnostic and can be applied to any RL algorithm. To validate these properties, we conduct experiments on neural architecture selection and step-size selection for different deep RL algorithms. In addition, we evaluate various model selection algorithms within our framework, highlighting the importance of data-driven selection. Our work has close connections to meta learning algorithms for adaptive hyperparameter selection in RL \cite{Elfwing2017OnlineMB, parker2022automated}, and addresses a key challenge of continual learning in non-stationary environments.  

\section{Preliminaries}

\subsection{Reinforcement Learning}
\label{subsec::perliminaries_RL}
We consider episodic reinforcement learning with a finite horizon $H$, which is formalized as a Markov Decision Process (MDP) $\langle \mathcal{S}, \mathcal{A}, \mathcal{R}, \mathcal{P}, \rho \rangle$. Here, $\mathcal{S}$ denotes the state space, $\mathcal{A}$ is the action space, $\mathcal{R}: \mathcal{S} \times \mathcal{A} \rightarrow \mathbb{R}$ is the reward function, $\mathcal{P}: \mathcal{S} \times \mathcal{A} \rightarrow [0,1]$ is the environment transition probabilities, and lastly $\rho: \mathcal{S} \rightarrow [0,1]$ is the initial state distribution. Denote $\pi: \mathcal{S} \rightarrow \Delta(\mathcal{A})$ as the policy of the agent. The agent interacts with the MDP according to the following procedure: At each step $h \in [H]$, the agent takes action $a_h \sim \pi(s_h)$, observes $r_h \sim \mathcal{R}(s_h, a_h)$ and move to state $s_{h+1} \sim \mathcal{P}(s_h, a_h)$. The value of the policy is defined as,
\begin{equation}
    \label{eq::value_function}
    v(\pi) = \mathbb{E} \left[\sum_{h=1}^H r_h \right]
\end{equation}
Here, the expectation is with respect to the stochasticity of the interaction procedure. The goal of the agent is to learn an optimal policy defined as,
\begin{equation}
    \label{eq::opt}
    \pi^* = \arg\max_{\pi \in \Pi} v(\pi)
\end{equation}

where $\Pi$ denotes the policy class. The policy class is commonly parameterized as $\Pi = \{ \pi_{\theta}: \theta \in \Theta \}$, where $\pi^* = \pi(\theta^*)$. The state value function $V: S \rightarrow \mathbb{R}$ and state-action value function $Q: S \times A \rightarrow \mathbb{R}$ with respect to policy $\pi$ are defined as,

\begin{equation}
    \label{eq::PG_update}
    V^{\pi}(s) = \mathbb{E} \biggl[\sum_{h=1}^{H}  r_h | s_0=s \biggr] 
\end{equation}
\begin{equation}
    \label{eq::QL-update}
    Q^{\pi}(s, a) = \mathbb{E} \biggl[\sum_{h=1}^{H}  r_h | s_0=s, a_0 = a\biggr]
\end{equation}

Note that definition \ref{eq::value_function} of the value function is equivalent to $v(\pi) = \mathbb{E}_{s \sim \rho}[V^{\pi}(s)]$, and $v(\pi) = \mathbb{E}_{s \sim \rho, \\ a \sim \pi}[Q^{\pi}(s, a)]$. From the algorithmic perspective, the agent does not try to directly solve the optimization problem in \ref{eq::opt}, as it can be computationally expensive, and in certain cases intractable. Instead, RL algorithm iteratively update the parameter $\theta$ to optimize the state-value function $V^{\pi}(s)$ or the state-action value function $ Q^{\pi}(s, a)$. Much of RL is devoted to designing algorithms with effective and sample-efficient update rules. But the most prominent themes are optimizing the advantage function, $ A^{\pi}(s, a) = Q^{\pi}(s, a) - V^{\pi}(s, a)$ through the policy gradient method \citep{williams1992simple, schulman2017proximal}, or optimizing the state-action value function by temporal difference methods \citep{watkins1992q, mnih2015human}, which we include in Appendix \ref{sec::appendix_rl_alg}. The details of the RL algorithm is not the focus this work, as we will show later that online model selection methods can be integrated into any RL training procedure.

\subsection{Model Selection}
\label{sec::preliminaries_model_selection}
We consider the online model selection problem where the selector has access to a set of  $M$ base agents, 
\begin{align*}
    \mathcal{B} =  \{\mathcal{B}^1, \ldots, \mathcal{B}^M \}
\end{align*}

At round $t \in [T]$, the selector picks base agent $i_t \in [M] $ according to its selection strategy. The selector rolls out the policy of the base agent $\pi^{i_t}_t$ for one episode and collects the trajectory,
\begin{align*}    
\tau = \{(s_h, a_h, r_h)\}_{h=1}^H  
\end{align*}
according to the procedure explained in \ref{subsec::perliminaries_RL}. The selector then uses this trajectory to update its selection policy, and forwards it to the base agents so that it can update its internal policy $\pi^i_t$.

\textbf{Notation:} Denote $n_t^i = \sum_{l=1}^t \mathbb{I}[i_l = i]$, as the number of rounds that base agent $\mathcal{B}^i$ has been selected up to round $t \in [T]$. Denote, $u_t^i = \sum_{l=1}^t \mathbb{I}[i_l = i]r_l$, where $r_l$ is the episodic reward at round $l \in [T]$, and $\bar{u}_t^i = \sum_{l=1}^t \mathbb{I}[i_l = i] \, \mathbb{E}[ r_l \mid \pi^i_l]$. The pseudo-regret of base agent $i \in [M]$ up to time $t \in [T]$, 
\begin{equation}
    \Regret_t^i = n_t^i v^* - \bar{u}_t^i
\end{equation}
where $v^* = \max_{\pi \in \Pi} v(\pi)$. The total regret of the selector after $T$ rounds is,
 \begin{align}
\Regret(T) = \sum_{i=1}^M \Regret_T^i = \sum_{i=1}^M n_T^i v^* - \bar{u}_T^i
\end{align}


The selector has access to base agents as sub-routines, and sequentially picks them in a meta-learning structure. 

\section{Training Mechanism}
\begin{figure}[t]
\centering
\begin{minipage}[t]{0.48\textwidth}
    \begin{algorithm}[H]
    \SetKwInput{KwInput}{Input}
    \DontPrintSemicolon
    
    \KwInput{$M$, $\mathcal{B}, \hat{d}^i_0 = d_{min} ~\forall i \in [M]$}
    
    \SetKwFunction{FSum}{sample}
    \SetKwFunction{FSub}{update}
    
    \SetKwProg{Fn}{Function}{:}{}
    \Fn{\FSum{}}{
        $i = \argmin_j \phi_t^j$\; 
        
        \KwRet $\mathcal{B}^{i}$\;
    } 
   
    \SetKwProg{Fn}{Function}{:}{}
    \Fn{\FSub{$i$, $r$, $t$}}{ 
        $u_{t+1}^i = u_t^i+r$, $\quad \quad n_{t+1}^i = n_t^i + 1$\;
        
        Perform Misspecification test \ref{eq::missp_test} for $\mathcal{B}^i$\;
        
        \If{Test Triggered}{
            $\hat{d}_t^i \leftarrow 2\hat{d}_t^i$\;
            
            $\phi_t^i = \hat{d}_t^i\sqrt{n_t^i}$\;
        }
    }
    \caption{Selector (D$^3$RB)}
    \end{algorithm}
\end{minipage}
\hfill
\begin{minipage}[t]{0.49\textwidth}
    \begin{algorithm}[H]
    \SetKwInput{KwInput}{Input}
    \DontPrintSemicolon
    
    \KwInput{Model Selector $selector$, $T$, $H$}
    \For{$t = 1, 2, ..., T$}{
        $i_t$ = $selector$.sample()\;
        
        \For{$h = 1, 2, ..., H$}{
            $a_h \sim \pi^{i_t}_{t}(\cdot| s_h)$\;
            
            $r_h \leftarrow \mathcal{R}(s_h,a_h) $\;
            
            $s_{h+1} \sim \mathcal{P}(s_h, a_h)$ \;
            
        }
        
        Forward $\{(s_h, a_h,r_h)\}_{h=1}^H$ to base agent $\mathcal{B}^{i_t}$ to update it policy
        
        $R_t = \sum_{h=1}^{H} r_h$\;
        
        $selector$.update($i_t,R_t, t$)\;
    }
    \caption{Training Mechanism}

    \end{algorithm}
\end{minipage}
\caption{Online Model Selection Framework in RL}
\label{alg::ModselRL}
\end{figure}

Online Model Selection in Reinforcement Learning deals with the problem of selecting over a set of \emph{evolving} policies. In this section, we characterize how the selector keeps track of the performance of base agents over time, and what is the right measure for that purpose. Importantly, we use this measure to specify the algorithmic objective of model selection.

\begin{definition}[Regret Coefficient]
    \label{def::realized_reg_coeff}
    For a positive constant $d_{min}$, denote the regret coefficient of base agent $\mathcal{B}^i \in \mathcal{B}$ at time $t\in[T]$ as,
    \begin{align}
        d_t^i  = \max \left( \frac{\Regret_t^i}{\sqrt{n_t^i}}, d_{min} \right)
    \end{align}
\end{definition}

\textbf{Algorithmic Objective in Model Selection} 
The goal in Model Selection is to design a meta-algorithm for the selector that has comparable performance to the base agent with minimum realized regret at the end of round $t \in [T]$. Define, 
\begin{align}
    d_* = \min_{i \in [M]} \max_{t \in [T]} d_t^i
\end{align}
as the regret coefficient of the well-specified base agent. The total regret incurred by the meta-algorithm should satisfy, 
\begin{align}
     \label{eq::model_objective}
     \Regret(T)  =   \sum_{i \in [M]}\Regret_T^i  \leq \text{Poly}(d_*) \sqrt{T}
\end{align}
with high probability.
\subsection{Algorithm}


Algorithm \ref{alg::ModselRL}-left shows the Doubling Data Driven Regret Balancing algorithm (D$^3$RB) \citep{dann2024data} that is the selector of interest in this paper. Algorithm \ref{alg::ModselRL}-right is the training mechanism, that shows how to integrate a selector into the RL training loop. The training mechanism is analogous to the standard agent-environment interaction protocol in RL with minimum interventions from the selector. As a result, this framework can be integrated into any RL algorithm. The choice of selector is also not limited to D$^3$RB, and other model selection algorithm can be used as long as they follow a similar interface to Algorithm \ref{alg::ModselRL}-left.  With that being said, the choice of selector matters in the performance and efficiency of the training mechanism. We choose D$^3$RB that satisfies the following high probability model selection guarantee,

\begin{theorem}[\citep{dann2024data}]
    \label{theroem::D3RB bound}
    Denote the event $\mathcal{E}$, 
    \begin{equation}
        \label{eq::good_event}
        \mathcal{E} = \left\{ \mid u_t^i - \bar{u}_t^i \mid \leq c \sqrt{ n_t^i \ln \frac{M \ln n_t^i}{\delta}} \right\}
    \end{equation}
    for a parameter $\delta$ and universal constant $c$. Under event $\mathcal{E}$,  the total regret incurred by data-driven regret balancing (D$^3$RB) after $T$ rounds satisfies,
    \begin{equation}
        \label{eq::D3RB regret bound}
        \Regret(T) = \mathcal{O} \left(d_* M\sqrt{T} + d_*^2 \sqrt{MT} \right)
    \end{equation}
    With probability $1-\delta$.
\end{theorem}  
This bound implies that D$^3$RB is optimal as it matches the lower bound of online model selection in \citep{marinov2021pareto, pacchiano2020model}.

D$^3$RB keeps track of a balancing potential function $\phi_t^i = \hat{d}_t^i \sqrt{n_t^i}$ for each base agent $i \in [M]$. This potential function estimates a \emph{data-adaptive} upper bound on the realized regret. Here, $\hat{d}_t^i$ is an active estimate of the true regret coefficient $d_t^i$.  At each round $t \in [T]$, the selector picks the base agent with minimum balancing potential $i_t = \argmin_{i \in [M]} \phi_t^i$. It acts according to the policy of the selected base agent $\pi_t^{i_t}$ for one episode, collects the trajectory $\tau = \{(a_h, s_h, r_h)\}_{h=1}^{H}$, and updates policy $\pi_t^{i_t}$ according to the update rule of the RL algorithm. The selector updates the statistics of the selected base agent, $n_t^{i_t}$, and $u_t^{i_t}$, and doubles $\hat{d}_t^{i_t}$ if the agent is misspecified.   

\begin{proposition}[Misspecification Test]
    Base agent $i \in [M]$ is misspecified if it triggers the following test,
    \begin{equation}
        \label{eq::missp_test}
        \frac{u_t^{i}}{n_t^{i}} +  c \sqrt{\ln \frac{\frac{M \ln n_t^{i}}{\delta}}{n_t^{i}}} +\frac{\hat{d}_t^{i} \sqrt{n_t^{i}}}{n_t^{i}} \leq \max_{j \in [M]} \frac{u_t^j}{n_t^j} - c \sqrt{\ln \frac{\frac{M \ln n_t^j}{\delta}}{n_t^j}}
    \end{equation}
\end{proposition}

We provide a detailed explanation in appendix \ref{sec::missp_test_exp} on why this test can determine misspecification of a base agent. In the following sections, we analyze the properties of the D$^3$RB+RL training mechanism in theory and demonstrate the effectiveness in experiments.

\section{Model Selection Tasks}
\label{sec::modsel_tasks}
\subsection{Neural Architecture Selection}
\label{sec::neural_architecture_selection}

\begin{figure}
    \centering

    \begin{subfigure}[b]{0.45\textwidth}
        \centering
        \includegraphics[width=\textwidth]{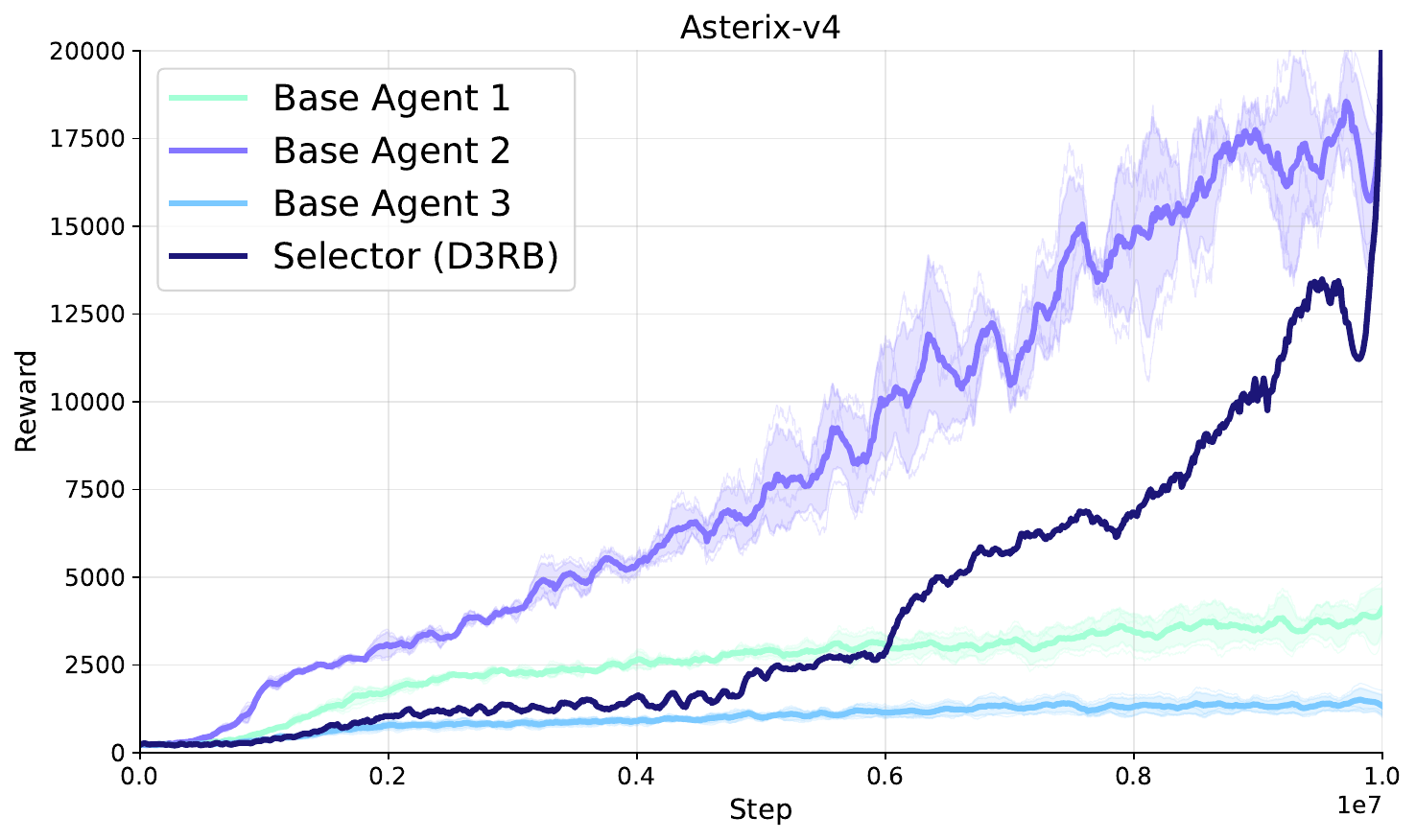}
        \caption{Asterix-v4}
        \label{fig:bottomleft}
    \end{subfigure}
    \hfill
    \begin{subfigure}[b]{0.45\textwidth}
        \centering
        \includegraphics[width=\textwidth]{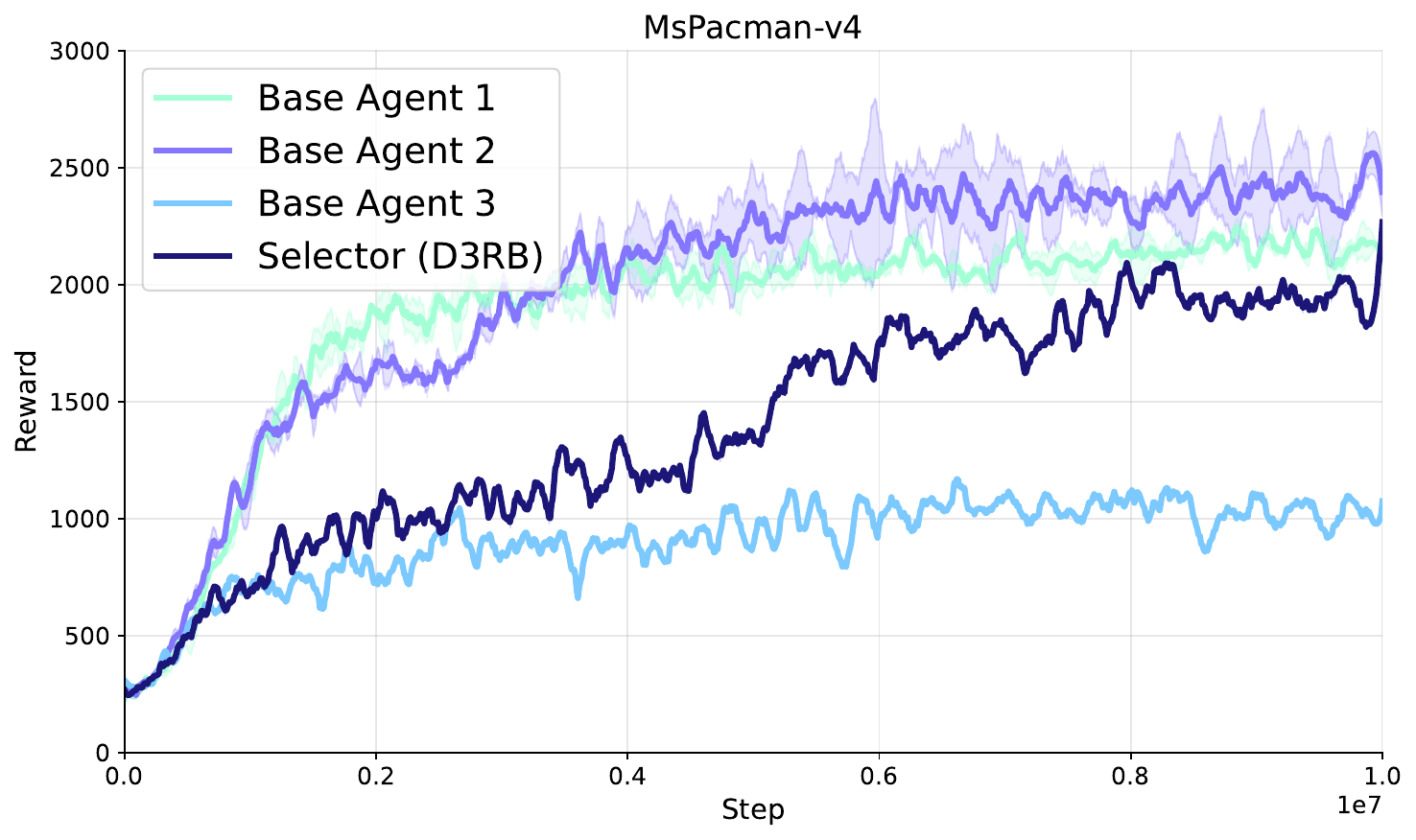}
        \caption{MsPacman-v4}
        \label{fig:topright}
    \end{subfigure}
    
    
    \begin{subfigure}[b]{0.45\textwidth}
        \centering
        \includegraphics[width=\textwidth]{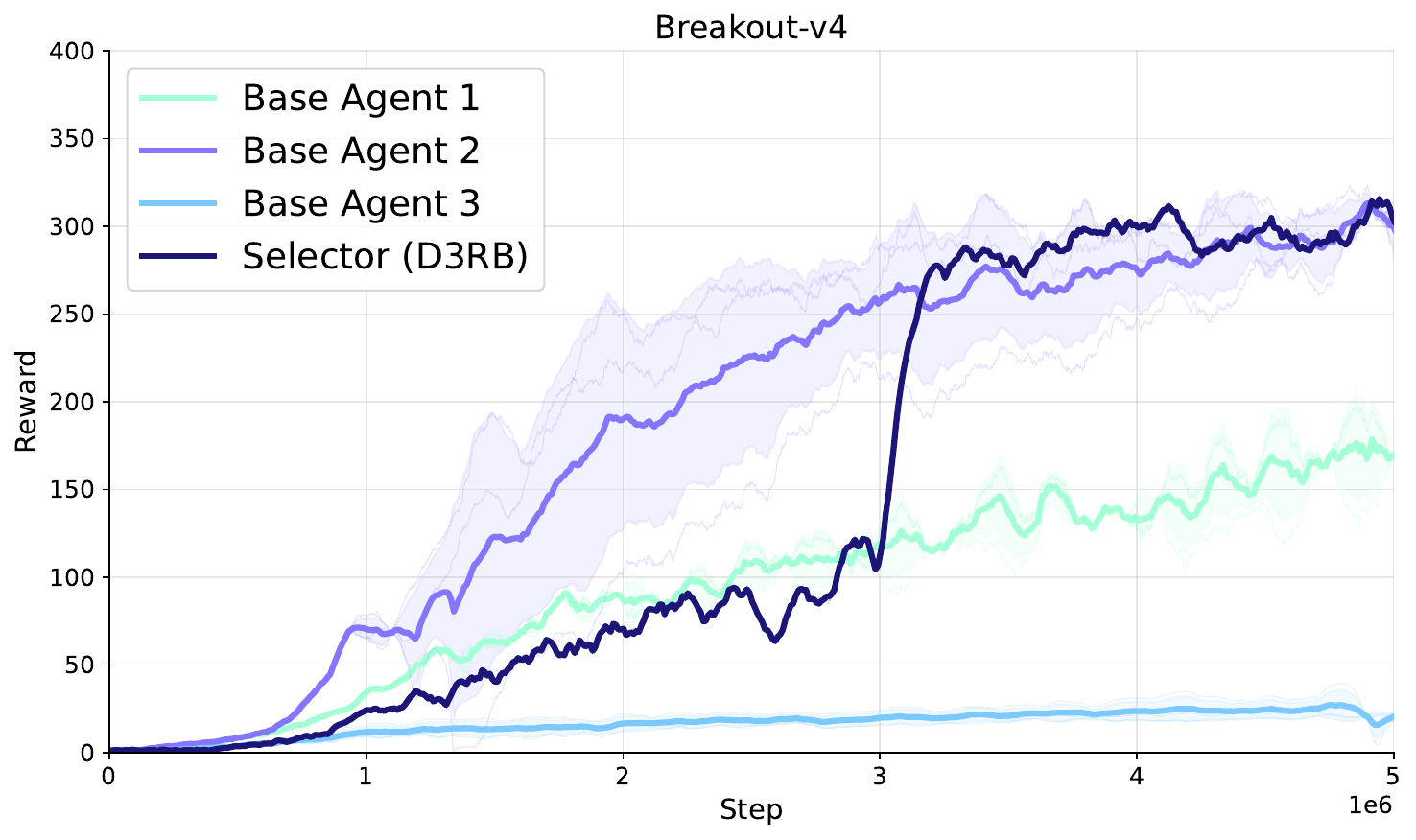}
        \caption{Breakout-v4}
        \label{fig:topleft}
    \end{subfigure}
    \hfill
    \begin{subfigure}[b]{0.45\textwidth}
        \centering
        \includegraphics[width=\textwidth]{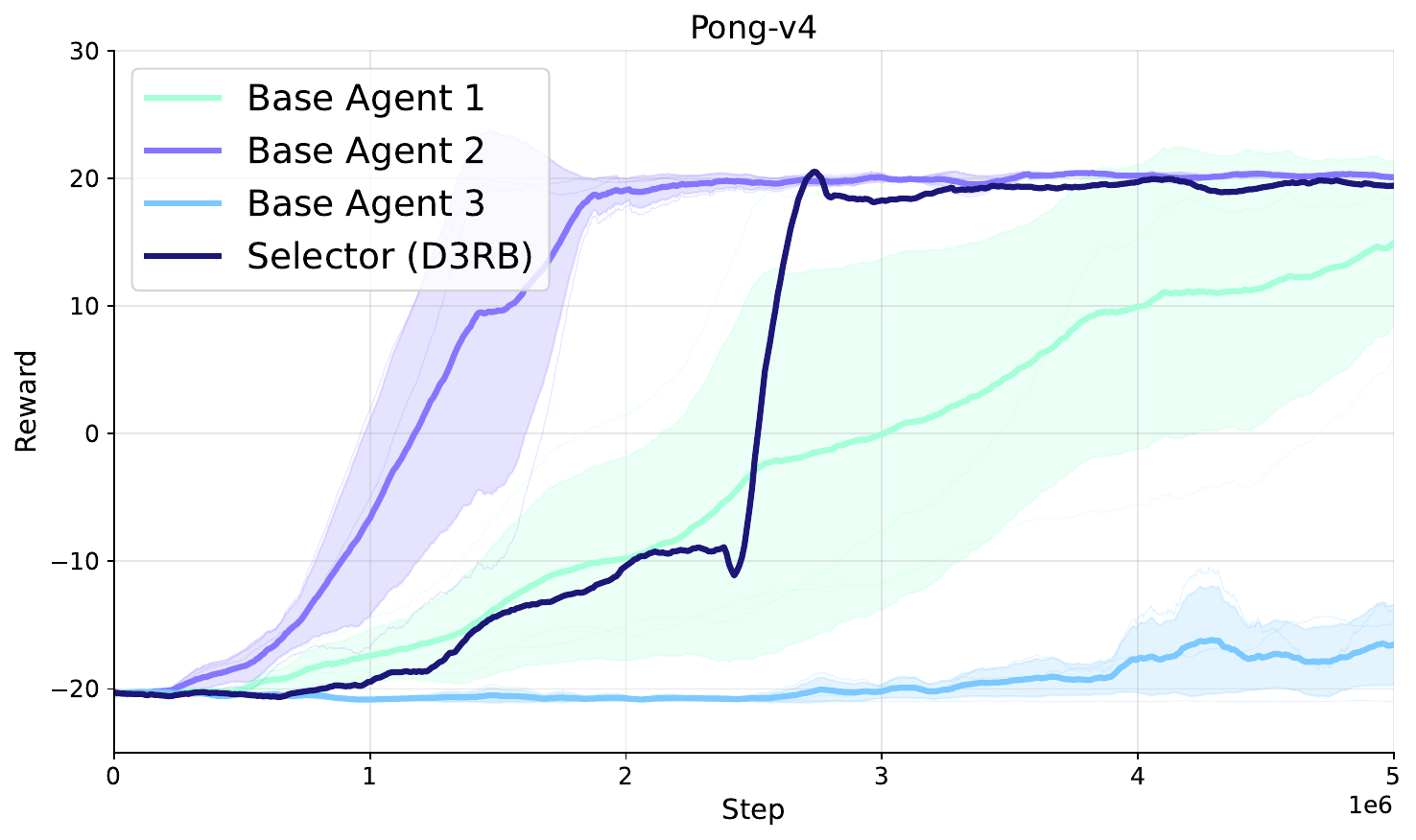}
        \caption{Pong-v4}
        \label{fig:bottomright}
    \end{subfigure}
    \caption{Neural Architecture Selection for DQN algorithm in Atari Environments. Comparison of D$^3$RB with individual runs of base agents shows that the selector has comparable performance to the best solo base agent. Curves show the average and standard deviation over three seeds.}
    \label{fig::Atari_Architecture_Figures}
\end{figure}

The policy in deep RL is parametrized by a neural network, which serves as an expressive function approximator, enabling decision making in high-dimensional state spaces that was otherwise not possible by linear function approximation or tabular RL. The choice of network architecture highly affects the agent's performance as it determines the representational capacity of the agent. 

\begin{figure}[h!]
\centering
\begin{tikzpicture}[node distance=0.3cm, every node/.style={scale=0.7}] 

\node (title0) [title] {Base Agent 1};
\node (conv0) [conv2, below=of title0] {Conv2d: 4→16};
\node (relu0) [relu, below=of conv0] {ReLU};
\node (flatten0) [flatten, below=of relu0] {Flatten};
\node (fc1_0) [fc3, below=of flatten0] {Linear: 6400→256};
\node (relu0b) [relu, below=of fc1_0] {ReLU};
\node (fc2_0) [fc, below=of relu0b] {Linear: 256→N};

\node (title1) [title, right=4cm of title0] {Base Agent 2};
\node (conv1) [conv2, below=of title1] {Conv2d: 4→32};
\node (relu1) [relu, below=of conv1] {ReLU};
\node (conv1b) [conv2, below=of relu1] {Conv2d: 32→64};
\node (relu1b) [relu, below=of conv1b] {ReLU};
\node (conv1c) [conv3, below=of relu1b] {Conv2d: 64→64};
\node (relu1c) [relu, below=of conv1c] {ReLU};
\node (flatten1) [flatten, below=of relu1c] {Flatten};
\node (fc1_1) [fc2, below=of flatten1] {Linear: 3136→512};
\node (relu1d) [relu, below=of fc1_1] {ReLU};
\node (fc2_1) [fc, below=of relu1d] {Linear: 512→N};

\node (title2) [title, right=4cm of title1] {Base Agent 3};
\node (conv2) [conv, below=of title2] {Conv2d: 4→8};
\node (relu2) [relu, below=of conv2] {ReLU};
\node (conv2b) [conv, below=of relu2] {Conv2d: 8→8};
\node (relu2b) [relu, below=of conv2b] {ReLU};
\node (conv2c) [conv, below=of relu2b] {Conv2d: 8→8};
\node (relu2c) [relu, below=of conv2c] {ReLU};
\node (flatten2) [flatten, below=of relu2c] {Flatten};
\node (fc1_2) [fc, below=of flatten2] {Linear: 392→32};
\node (relu2d) [relu, below=of fc1_2] {ReLU};
\node (fc2_2) [fc, below=of relu2d] {Linear: 32→N};

\foreach \i/\j in {conv0/relu0, relu0/flatten0, flatten0/fc1_0, fc1_0/relu0b, relu0b/fc2_0,
                    conv1/relu1, relu1/conv1b, conv1b/relu1b, relu1b/conv1c, conv1c/relu1c, relu1c/flatten1, flatten1/fc1_1, fc1_1/relu1d, relu1d/fc2_1,
                    conv2/relu2, relu2/conv2b, conv2b/relu2b, relu2b/conv2c, conv2c/relu2c, relu2c/flatten2, flatten2/fc1_2, fc1_2/relu2d, relu2d/fc2_2}
    \draw[->] (\i) -- (\j);

\end{tikzpicture}
\caption{Q-Network architecture for three base agents in neural architecture selection task}
\label{fig::base_architectures}
\end{figure}
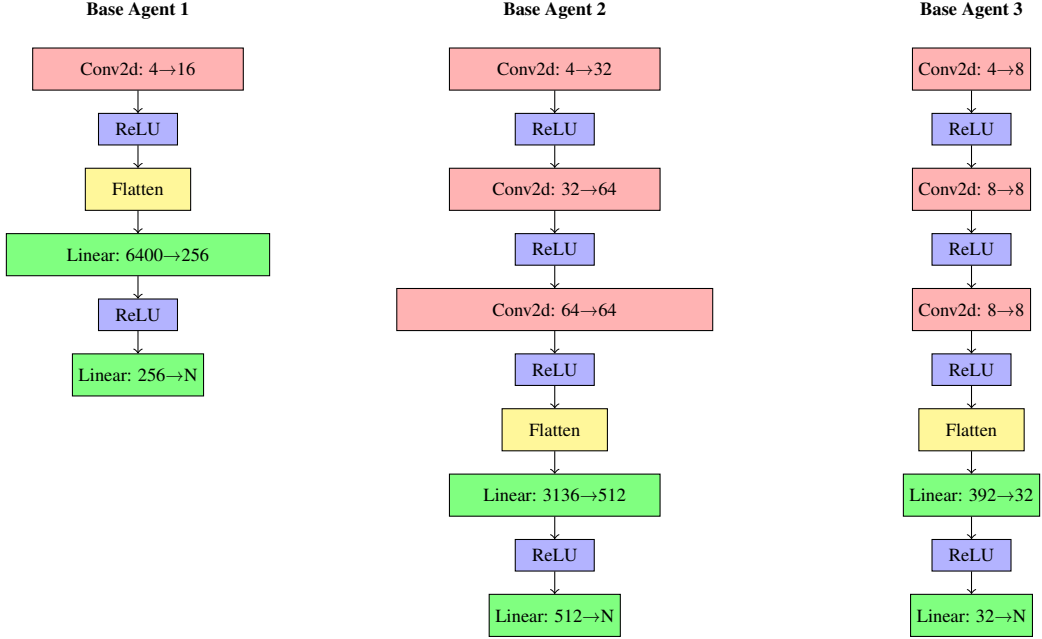

To validate the practical efficiency of our mechanism, we run experiments for the neural architecture selection task in DQN agents for 4 different Atari environments \cite{mnih2015human}. We pair the D$^3$RB selector with three base agents that are identified with their unique Q-network architecture. Figure \ref{fig::Atari_Architecture_Figures} and \ref{fig::base_architectures} depict the results of the neural architecture selection task and the Q-network architecture of each base agent respectively. To better see the progressive performance of the selector, we also included the independent execution of base agents. We observe that the performance of base agents varies drastically depending on the architectural choice, as base agent $\mathcal{B}^3$ is failing in almost all of the environments, $\mathcal{B}^1$ has sub-optimal performance, and $\mathcal{B}^2$ is the oracle-best. Importantly, the reward curve of the model selection approach shows that the D$^3$RB selector is able to reach the performance of the oracle-best base agent, confirming the theoretical objective of model selection \ref{eq::model_objective} in practice. In the following theorem, we derive the relationship between the allocated compute and regret coefficient of each base learner under Data-Driven Model Selection.

\begin{theorem}
    \label{theorem:: D3RB resource allocation}
    Denote $\alpha_t^i$ as the fraction of time that base agent $\mathcal{B}^i \in \mathcal{B}$ has been selected up to round $t$, and $d_t^i$ as the regret coefficient \ref{def::realized_reg_coeff} of $\mathcal{B}^i$ at time $t$,
    \begin{align*}
        \alpha_t^i = \frac{\sum_{l=1}^t \mathbb{I}[i_l=i]}{t} , \quad\quad  \forall i \in[M],  t\in [T]
    \end{align*}
    then, data-driven model selection (D$^3$RB) satisfies, 
    \begin{align*}
        \alpha_t^i = \frac{(1/d_t^i)^2}{\sum_{j=1}^M (1/d_t^j)^2}
    \end{align*}
    
    \proof Appendix \ref{sec::proof_D3RB_resource_allocation}
\end{theorem}

Theorem \ref{theorem:: D3RB resource allocation} shows that for a fixed training budget $T$, the base agent with a smaller regret coefficient (and hence a better realized performance) will receive a bigger fraction of compute under model selection with D$^3$RB. Table \ref{table::architecture_selection_statistics} summarizes the resource allocation of neural architecture selection experiments \ref{fig::Atari_Architecture_Figures}. For each environment, we calculated the fraction of rounds in which a base agent was selected throughout training. These results show that D$^3$RB has learned to allocate more compute to the oracle-best base agent ($\mathcal{B}^2$). The oracle-best base agent is determined by the independent execution of base agents.

\begin{table}[h]
\centering
\begin{tabular}{l|c|c|c}

 Environment & \textbf{$\mathcal{B}^1$} & \textbf{$\mathcal{B}^2$} & \textbf{$\mathcal{B}^3$} \\
 &  & (Oracle-Best) &  \\
\hline
Breakout-v4 & 0.29 &  \textbf{0.43} & 0.28 \\
\hline
MsPacman-v4 & 0.40 &  \textbf{0.41} & 0.19 \\
\hline
Asterix-v4 & 0.29 &  \textbf{0.46} & 0.25 \\
\hline
Pong-v4 & 0.21 &  \textbf{0.58} & 0.21 \\

\end{tabular}
\caption{Resource allocation in neural architecture selection task under D$^3$RB. Each number shows the fraction of time that D$^3$RB has selected each base agent during training. The maximum amount of selection is in bold.}
\label{table::architecture_selection_statistics}.
\end{table}

\subsection{Step-Size Selection}
\label{sec::learning_rate_selection}

In this section, we consider the step size selection task in RL, where we are interested in adaptively selecting the right step-size for a given RL problem. We use this task to answer why bandit algorithms are insufficient to perform model selection for RL agents and compare data-driven model selection methods with other baselines.

\subsubsection{Bandit Algorithms for Online Model Selection: What would go wrong?}
As an algorithmic counterpart to online model selection, consider the problem of Multi-Armed Bandits (MAB)\citep{lattimore2020bandit} where the learner interacts with a finite set of arms. One can consider to use a MAB algorithm for the model selection problem, where each arm is a base learner and the MAB algorithm acts as a selector. We argue that standard Bandit algorithms with regret-minimization objective are insufficient to perform model selection for RL agents.
\begin{enumerate}
    \item The standard assumption in MAB is that each arm has an unknown but fixed reward distribution, whereas in model selection for RL agents, $\mathbb{E}[r \mid \pi_t^i]$ changes as policy gets updated over time.
    \item The optimal arm in standard MAB is associated with a fixed arm index. In model selection, the optimal base agent can change in different stages of training. As an example, consider two base agents that one initially performs worse than the other, but outperforms the other base agent towards the end of training. Ideally, the model selection strategy should take this non-stationarity into account and adapt to the new optimal. 
\end{enumerate}

\begin{figure}
    \centering
    \begin{subfigure}[b]{0.32\textwidth}
        \centering
        \includegraphics[width=\textwidth]{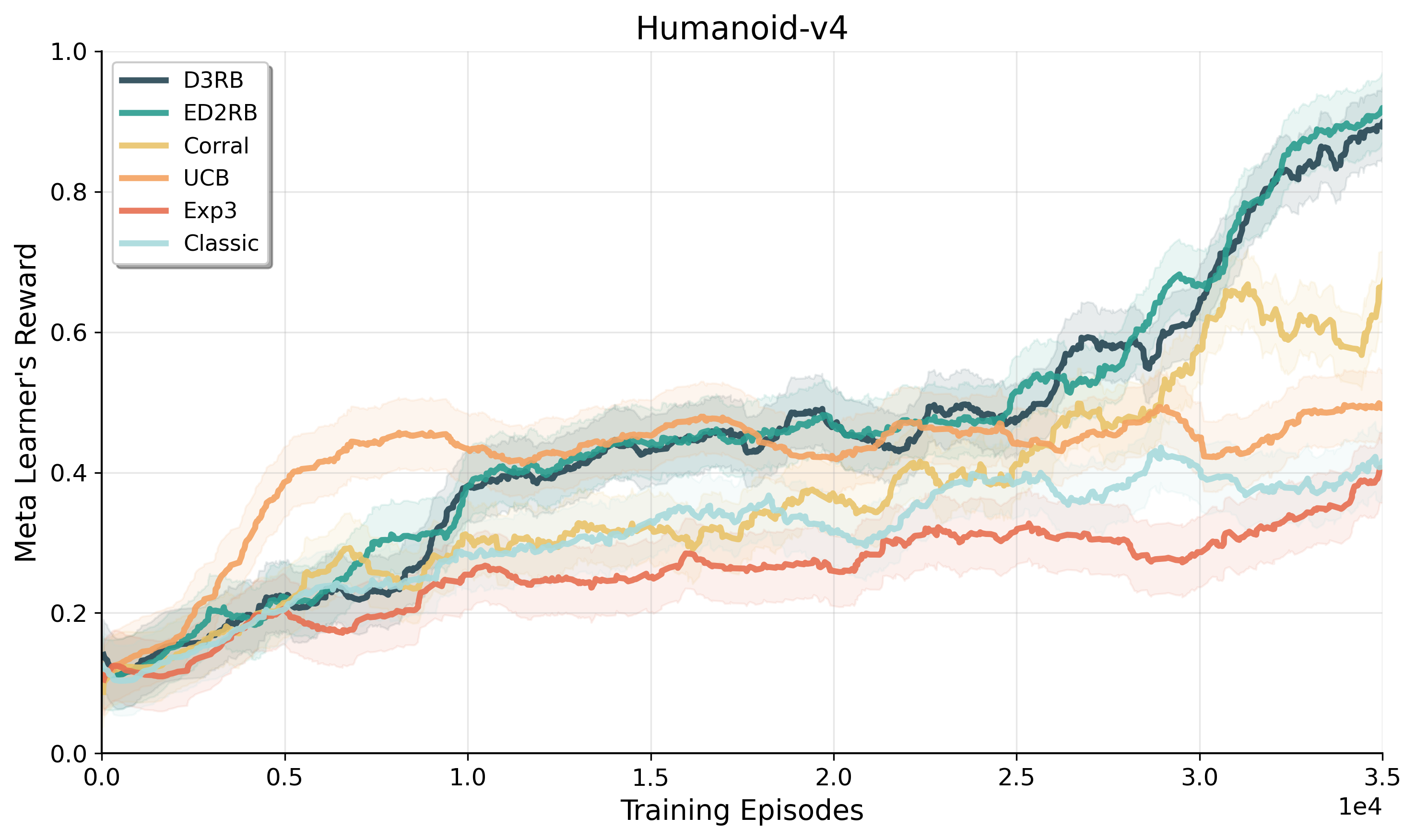}
        \caption{Humanoid-v4}
        \label{fig:left}
    \end{subfigure}
    \hfill
    \begin{subfigure}[b]{0.32\textwidth}
        \centering
        \includegraphics[width=\textwidth]{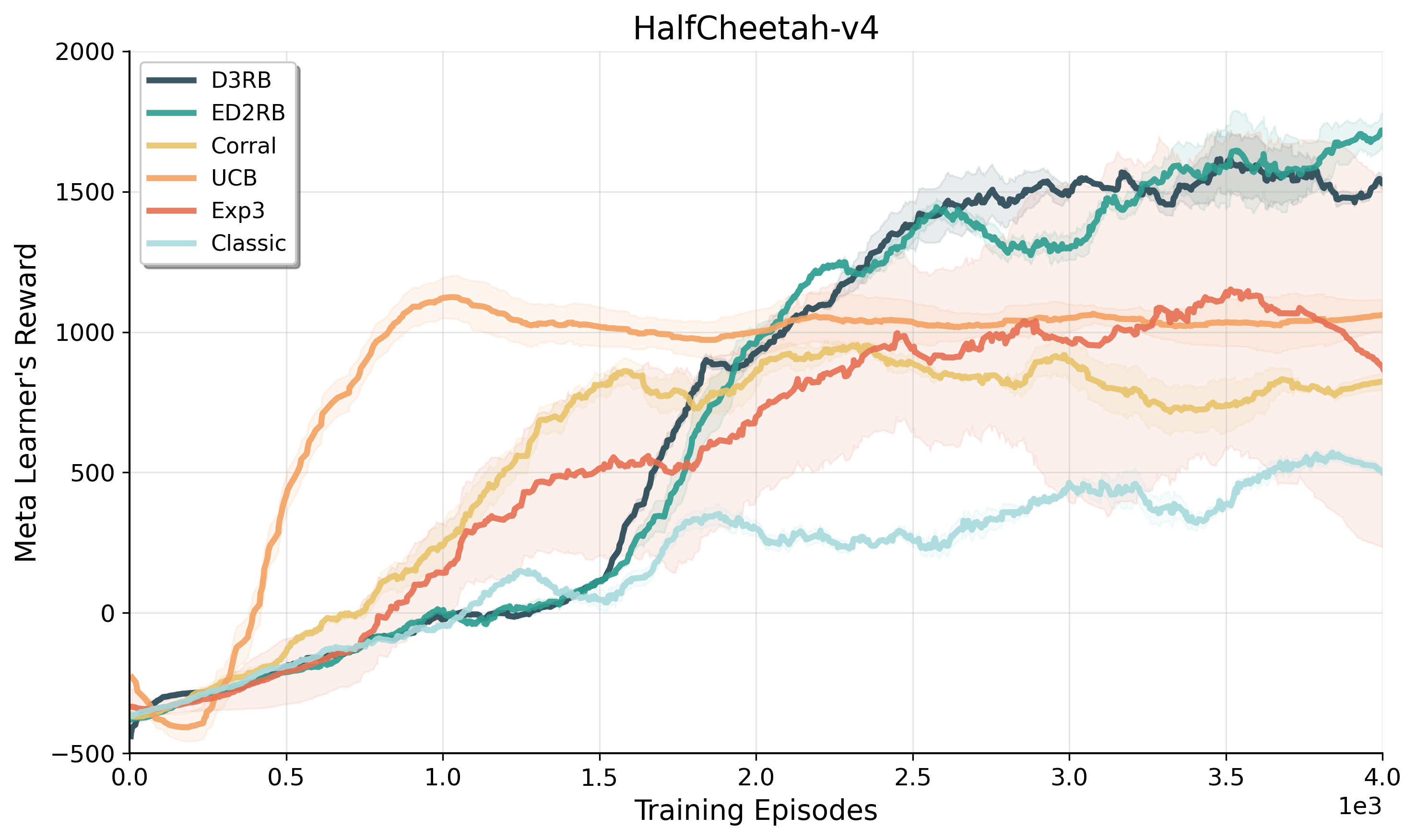}
        \caption{HalfCheetah-v4}
        \label{fig:center}
    \end{subfigure}
    \hfill
    \begin{subfigure}[b]{0.32\textwidth}
        \centering
        \includegraphics[width=\textwidth]{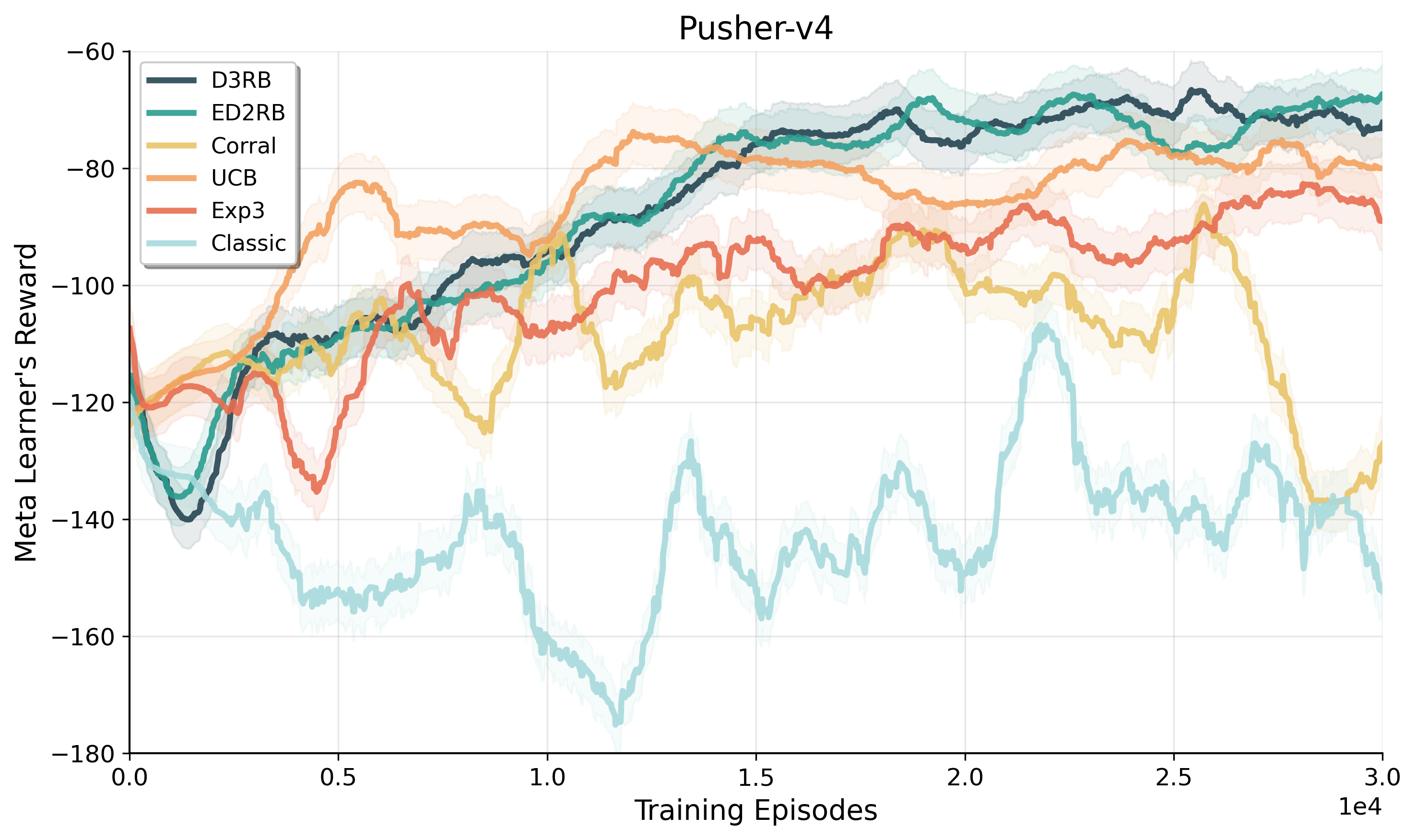}
        \caption{Pusher-v4}
        \label{fig:right}
    \end{subfigure}
    
    \caption{Comparison of 6 model selection strategies in the step-size selection task for the PPO algorithm. Each curve shows the average and standard deviation over three seeds.}
    \label{fig::step_size_selection_baselines}
\end{figure}

The results in the figure \ref{fig::step_size_selection_baselines} empirically validate the argument above. In these experiments, we consider the step size selection for PPO agents in three different MuJoco environments \citep{6386025}. We initiate five PPO base agents with the celebrated logarithmic scale values for hyperparameter tuning $[1e^{-2}, 1e^{-3}, 1e^{-4}, 1e^{-5}, 1e^{-6}]$. We pair these base agents with six different model selection algorithms. The model selection strategies are D$^3$RB, ED$^2$EB \citep{dann2024data}, Corral \citep{agarwal2017corralling}, Regret Bound Balancing \cite{pacchiano2020regretboundbalancingelimination}, which we will refer to as Classic Balancing. We consider two standard Bandit algorithms, Upper Confidence Bound (UCB) \citep{auer2002finite}, and Exponential-weight algorithm for Exploration and Exploitation (EXP3) \citep{bubeck2012regret}.

\begin{figure}[htbp]
    \centering
    \includegraphics[width=0.3\textwidth]{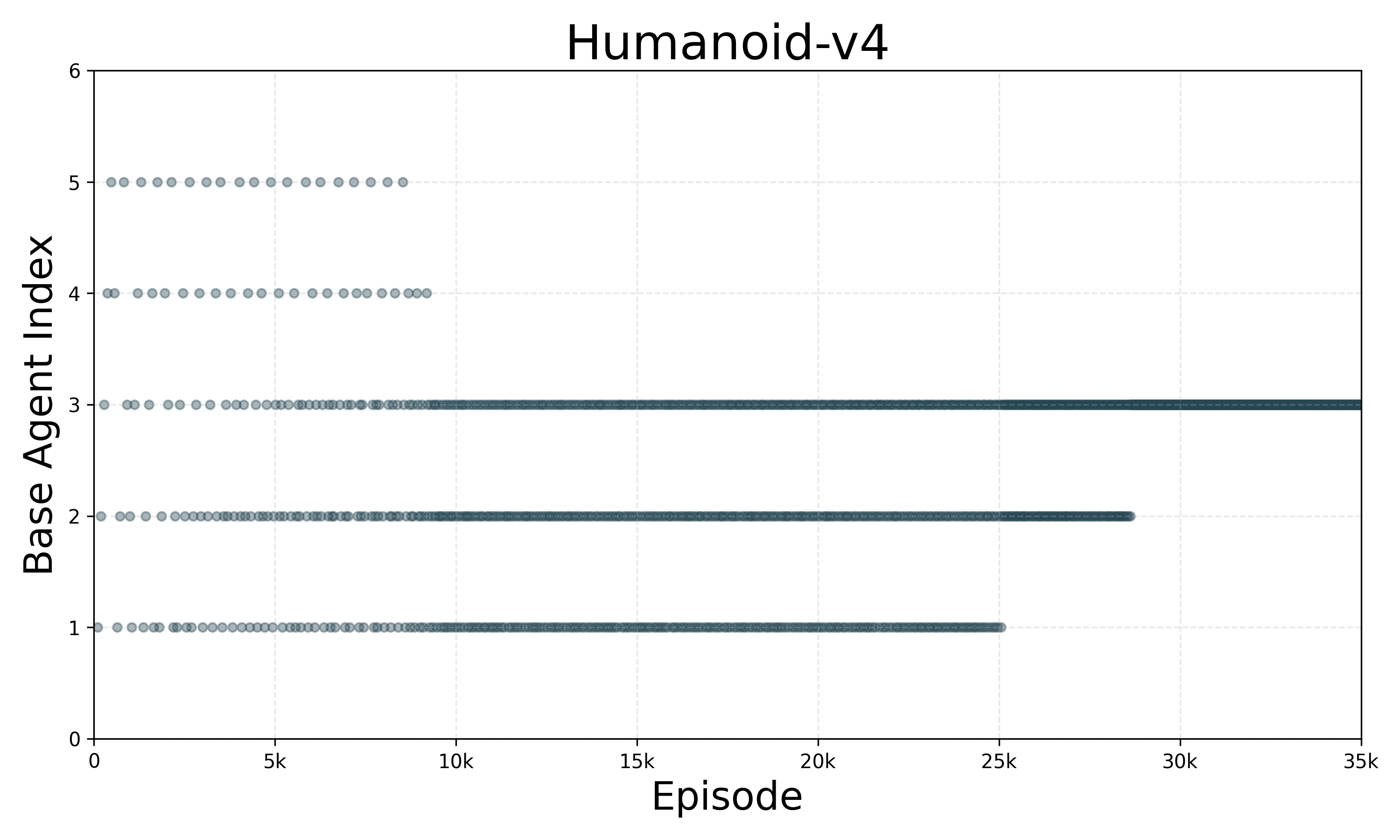}
    \hfill
    \includegraphics[width=0.3\textwidth]{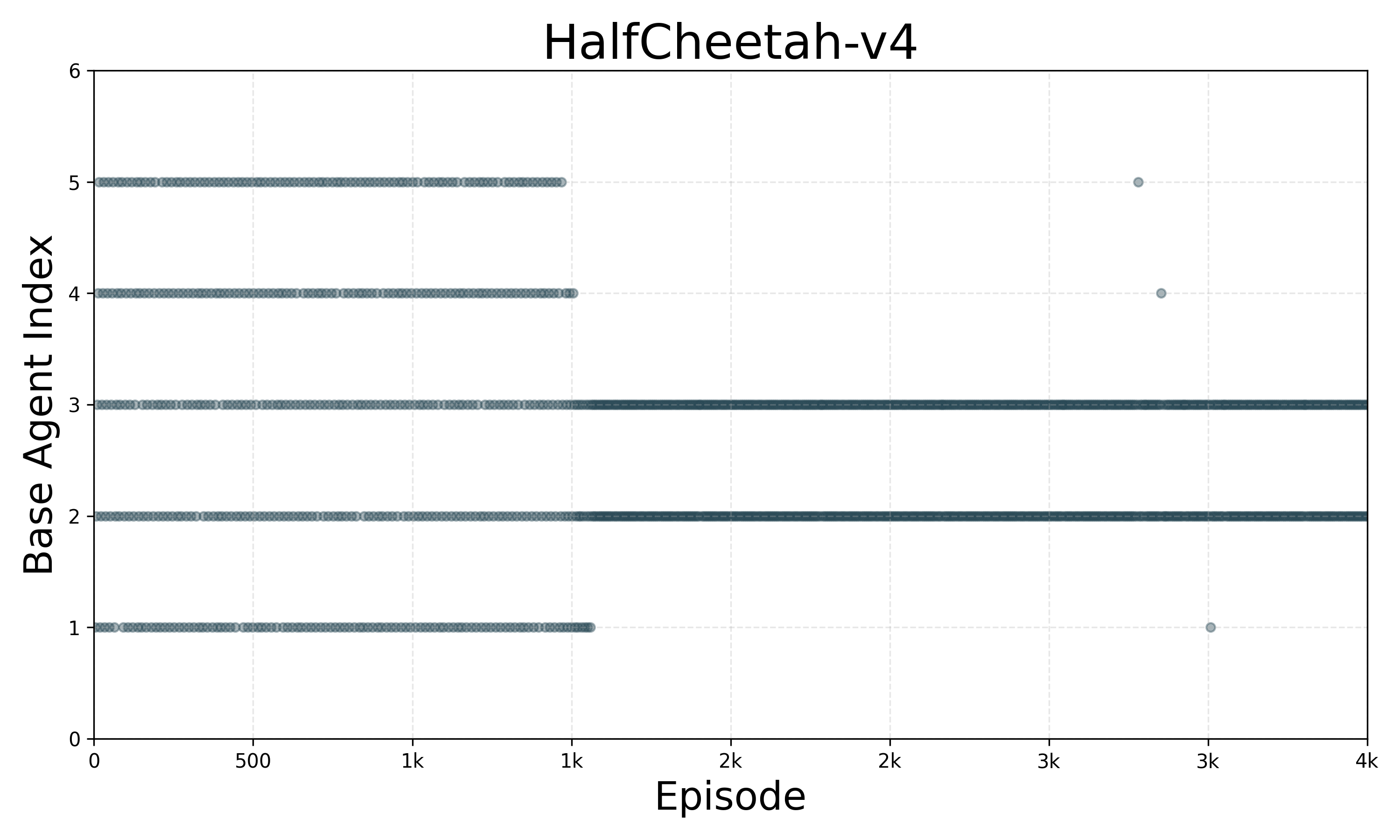}
    \hfill
    \includegraphics[width=0.3\textwidth]{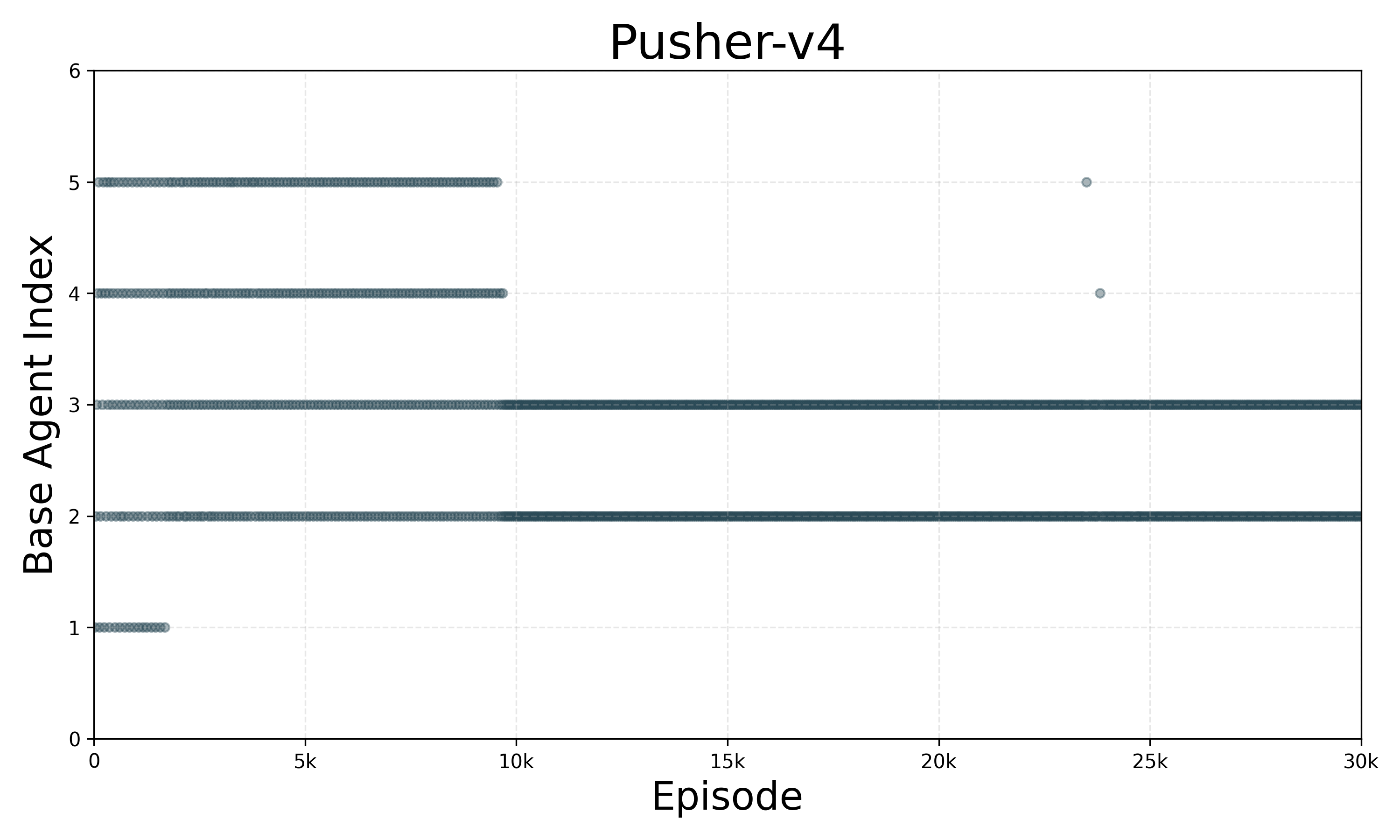}
    \caption{D$^3$RB Selection Statistics in MuJoco Environments}
    \label{fig::D3RB selection}
\end{figure}

\begin{figure}[htbp]
    \centering
    \includegraphics[width=0.3\textwidth]{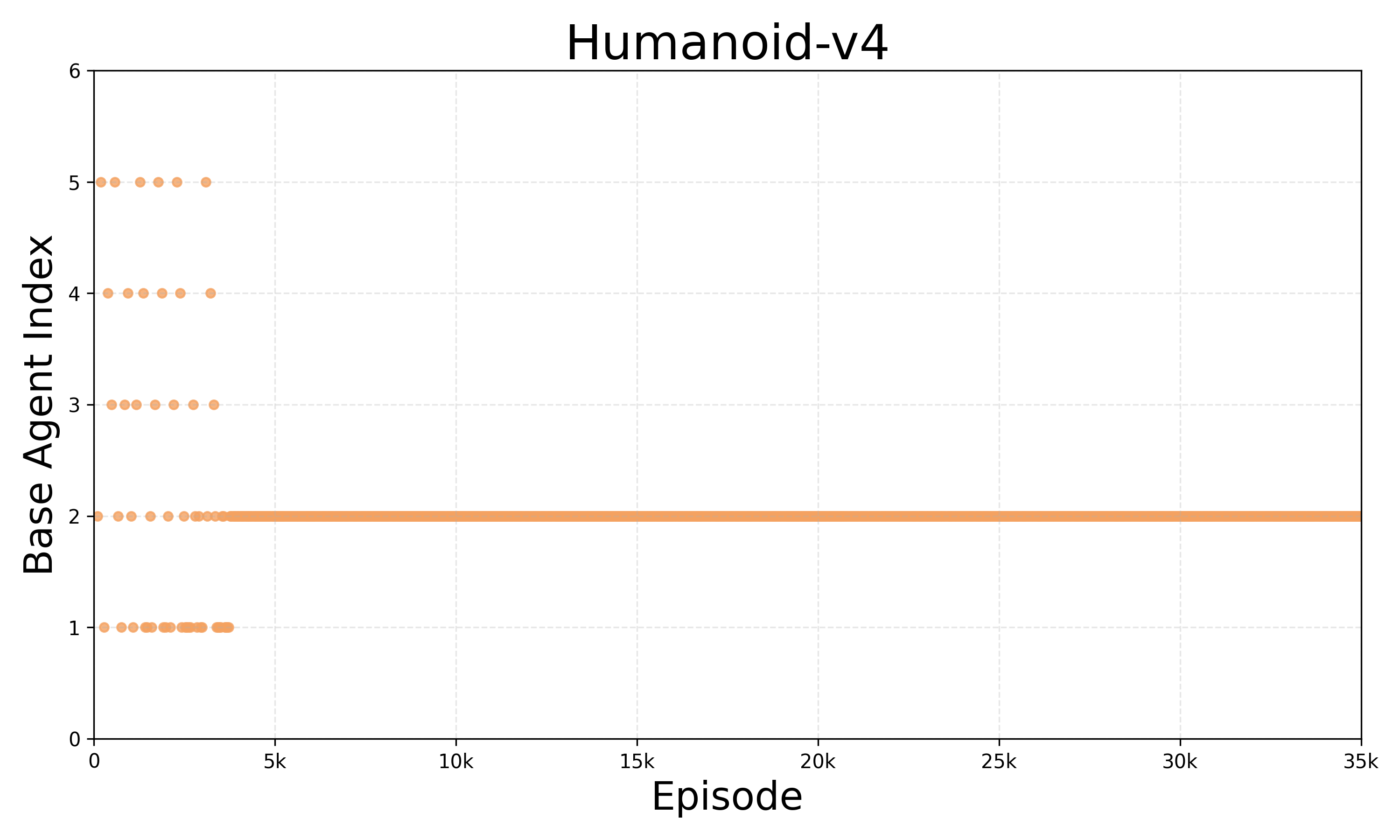}
    \hfill
    \includegraphics[width=0.3\textwidth]{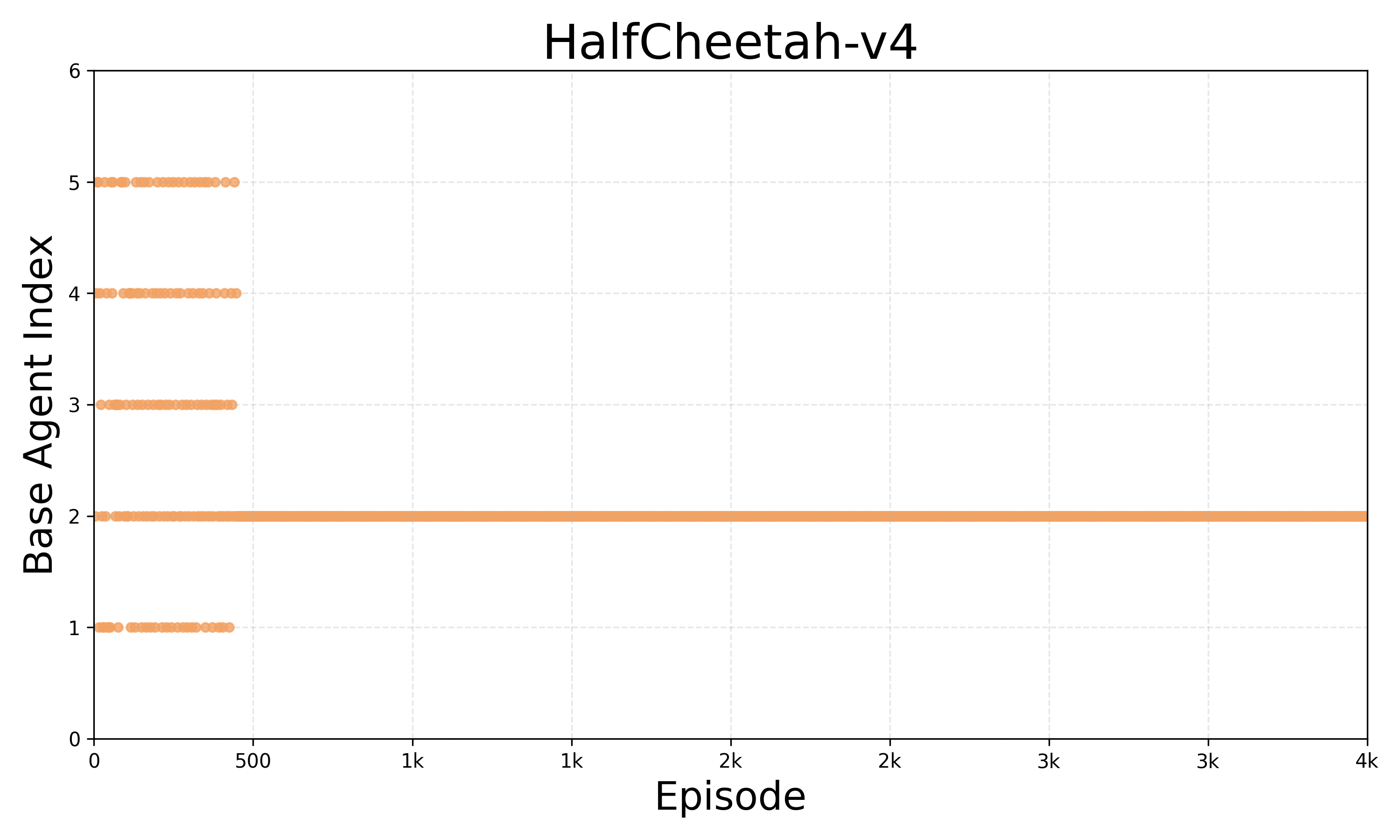}
    \hfill
    \includegraphics[width=0.3\textwidth]{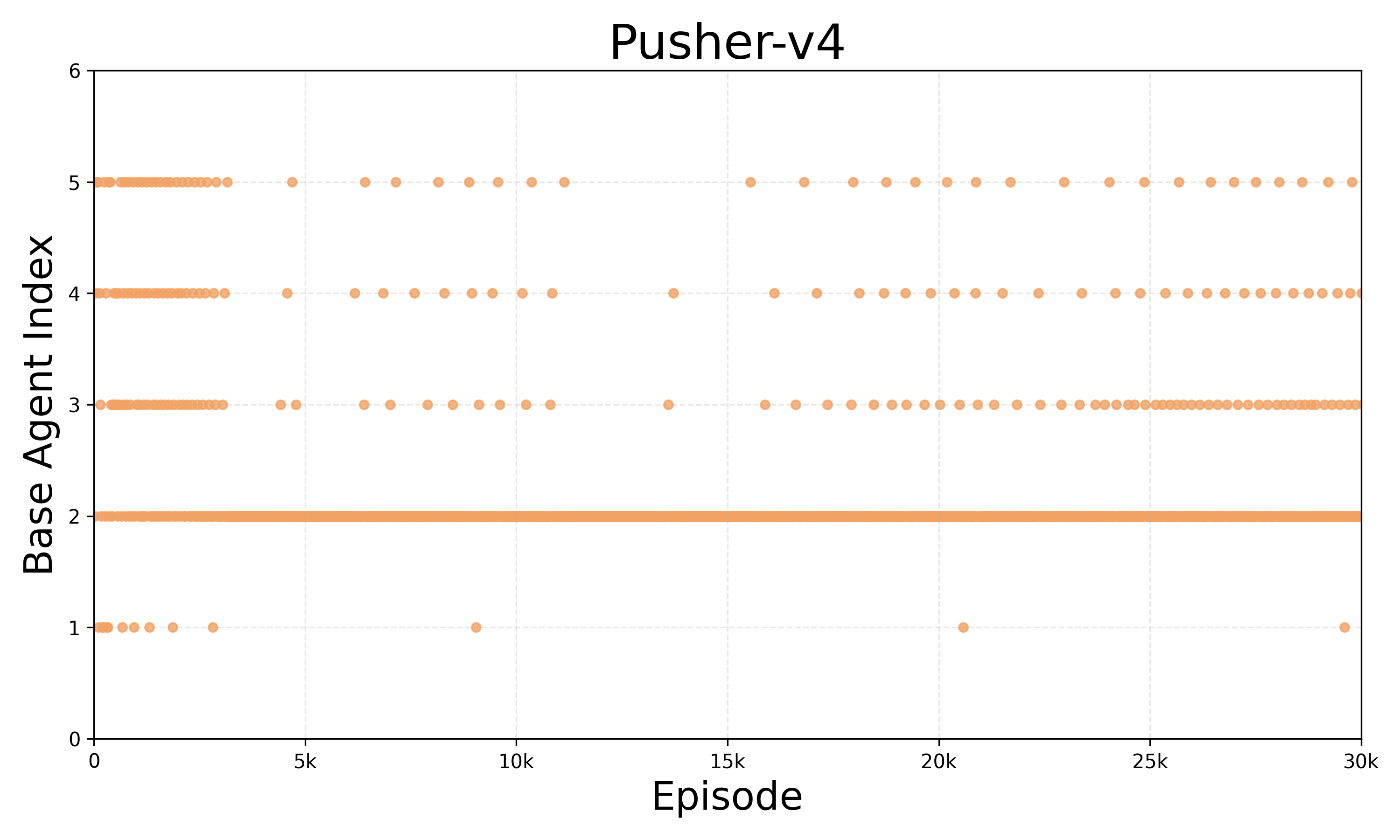}
    \caption{UCB Selection Statistics in MuJoco Environments}
    \label{fig::UCB_selection}
\end{figure}

The reward curves demonstrate that the choice of selector matters in the model selection task, as D$^3$RB and ED$^2$RB (which have similar model selection guarantees) achieve superior performance in comparison to other model selection strategies. Additionally, the reward curve of UCB highlights that the standard assumptions of Bandits might fail in RL model selection tasks. We observe that UCB is able to dominate other selectors at the initial phase of training, but the performance plateaus towards the end. We explain this behavior by investigating the detailed selection statistics of the model selection algorithm. Figure \ref{fig::D3RB selection} illustrates how base agents are picked by selectors throughout the training. We observe that UCB is overcommitting to a base agent that was optimal at the initial stage of training, but fails to remain explorative and adapt to new step sizes that are optimal towards the end. We include the selection statistics of all algorithms in Appendix \ref{sec::step_size_selection_plots}. We further analyze the results of the Classic algorithm and the effect of misspecification test in Appendix \ref{subsec::perliminaries_RL}.

\subsection{Self-Model Selection}
\label{sec::self_model_selection}

Self-model selection refers to the setting where base agents have identical configurations but are executed with different initial randomization (seed). This setting can stabilize the training of RL algorithms whose performance is highly sensitive to the seeding. We show that by combining certain number of base agents that each fail with a fixed probability, self model selection can capitalize on the successful runs and achieve its bound. The following theorem formalizes this property, 



\begin{theorem}
    \label{theorem::self-model-selection}
    Let $\delta \in (0,1)$, and $\mathcal{R}_\star(t, \delta): ([T], (0,1)) \rightarrow\mathbb{R}_+$ be a function satisfying 
    \begin{align}
        \mathcal{R}_\star(t,\delta) \leq d_\star \sqrt{t} \quad\quad \forall t \in [T]
    \end{align}
    Suppose a learning agent $\mathcal{B}$ satisfies,
    \begin{equation}\label{equation::constant_probability_success}    \mathbb{P}\left(\Regret_t^\mathcal{B} \leq \mathcal{R}_\star(t, \delta) \quad \forall t \in [T]  \right) \geq 1 - \gamma(\delta)
    \end{equation}
    for some $\gamma(\delta) \in (0,1) $. Then, for $M = \lceil\frac{\log\left(\delta\right)}{\log (\gamma(\delta))} \rceil$, D$^3$RB achieves the bound,
    \begin{equation*}
        \Regret(T) = \mathcal{O} \left(d_* M\sqrt{T} + d_*^2 \sqrt{MT} \right)
    \end{equation*}
    with probability at least $1-\delta$. 
    \proof Appendix  \ref{sec::proof_self_model_selection}
\end{theorem}






\section{Discussion and Future Work}
The theoretical guarantee \ref{eq::D3RB regret bound} of data-driven model selection reflects that the regret of the selector grows linearly with the number of base agents $M$. This is a barrier for deploying these methods to large-scale model selection tasks and incentivizes designing methods with improved dependency on $M$. Prior work of \citep{kassraie2024anytime} has studied this problem for the case of linear bandits, but the question remains open for RL. 

One can consider a similar model selection framework to ours, where base agents can communicate or share data. From the algorithmic perspective, a natural idea is to use importance sampling to update the policy of one base agent with the trajectory collected by another base agent. We discuss this more in details in Appendix \ref{sec::sharing_data}, but we leave the theoretical analysis of this setup as future work. 

Apart from the model selection tasks studied in this paper, prior work has studied online reward selection \citep{zhang2024orso} in policy optimization methods. One can consider other practical model selection tasks, such as selection of exploration strategy or selection of pretrained policy.

\section{Conclusion}

This work presents a principled mechanism to improve the efficiency of RL training procedure via online model selection. We studied data-driven model selection methods, how they characterize misspecification of base agents, and their theoretical guarantees. We showed, through theoretical analysis and empirical evaluation, that these methods 1) learn to adaptively direct more compute towards the base agents with better realized performance, 2) adapt to new optimal choices under non-stationary dynamics, and 3) enhance training stability. In addition, we validated these properties on a range of practical model selection tasks in deep reinforcement learning.

\newpage

\bibliography{iclr2026_conference}

\begin{thebibliography}{25}
\providecommand{\natexlab}[1]{#1}
\providecommand{\url}[1]{\texttt{#1}}
\expandafter\ifx\csname urlstyle\endcsname\relax
  \providecommand{\doi}[1]{doi: #1}\else
  \providecommand{\doi}{doi: \begingroup \urlstyle{rm}\Url}\fi

\bibitem[Agarwal et~al.(2017)Agarwal, Luo, Neyshabur, and Schapire]{agarwal2017corralling}
Alekh Agarwal, Haipeng Luo, Behnam Neyshabur, and Robert~E Schapire.
\newblock Corralling a band of bandit algorithms.
\newblock In \emph{Conference on Learning Theory}, pages 12--38. PMLR, 2017.

\bibitem[Agarwal et~al.(2019)Agarwal, Jiang, and Kakade]{agarwal2019reinforcement}
Alekh Agarwal, Nan Jiang, and Sham~M Kakade.
\newblock Reinforcement learning: Theory and algorithms.
\newblock 2019.

\bibitem[Auer et~al.(2002)Auer, Cesa-Bianchi, and Fischer]{auer2002finite}
Peter Auer, Nicolo Cesa-Bianchi, and Paul Fischer.
\newblock Finite-time analysis of the multiarmed bandit problem.
\newblock \emph{Machine learning}, 47:\penalty0 235--256, 2002.

\bibitem[Bubeck et~al.(2012)Bubeck, Cesa-Bianchi, et~al.]{bubeck2012regret}
S{\'e}bastien Bubeck, Nicolo Cesa-Bianchi, et~al.
\newblock Regret analysis of stochastic and nonstochastic multi-armed bandit problems.
\newblock \emph{Foundations and Trends{\textregistered} in Machine Learning}, 5\penalty0 (1):\penalty0 1--122, 2012.

\bibitem[Cutkosky et~al.(2021)Cutkosky, Dann, Das, Gentile, Pacchiano, and Purohit]{cutkosky2021dynamic}
Ashok Cutkosky, Christoph Dann, Abhimanyu Das, Claudio Gentile, Aldo Pacchiano, and Manish Purohit.
\newblock Dynamic balancing for model selection in bandits and rl.
\newblock In \emph{International Conference on Machine Learning}, pages 2276--2285. PMLR, 2021.

\bibitem[Dann et~al.(2024)Dann, Gentile, and Pacchiano]{dann2024data}
Chris Dann, Claudio Gentile, and Aldo Pacchiano.
\newblock Data-driven online model selection with regret guarantees.
\newblock In \emph{International Conference on Artificial Intelligence and Statistics}, pages 1531--1539. PMLR, 2024.

\bibitem[Elfwing et~al.(2017)Elfwing, Uchibe, and Doya]{Elfwing2017OnlineMB}
Stefan Elfwing, Eiji Uchibe, and Kenji Doya.
\newblock Online meta-learning by parallel algorithm competition.
\newblock \emph{Proceedings of the Genetic and Evolutionary Computation Conference}, 2017.
\newblock URL \url{https://api.semanticscholar.org/CorpusID:8582140}.

\bibitem[Foster and Rakhlin(2023)]{foster2023foundations}
Dylan~J Foster and Alexander Rakhlin.
\newblock Foundations of reinforcement learning and interactive decision making.
\newblock \emph{arXiv preprint arXiv:2312.16730}, 2023.

\bibitem[Foster et~al.(2020)Foster, Gentile, Mohri, and Zimmert]{foster2020adapting}
Dylan~J Foster, Claudio Gentile, Mehryar Mohri, and Julian Zimmert.
\newblock Adapting to misspecification in contextual bandits.
\newblock \emph{Advances in Neural Information Processing Systems}, 33:\penalty0 11478--11489, 2020.

\bibitem[Kassraie et~al.(2024)Kassraie, Emmenegger, Krause, and Pacchiano]{kassraie2024anytime}
Parnian Kassraie, Nicolas Emmenegger, Andreas Krause, and Aldo Pacchiano.
\newblock Anytime model selection in linear bandits.
\newblock \emph{Advances in Neural Information Processing Systems}, 36, 2024.

\bibitem[Lattimore and Szepesv{\'a}ri(2020)]{lattimore2020bandit}
Tor Lattimore and Csaba Szepesv{\'a}ri.
\newblock \emph{Bandit algorithms}.
\newblock Cambridge University Press, 2020.

\bibitem[Lee et~al.(2022)Lee, Tucker, Nachum, Dai, and Brunskill]{lee2022oracle}
Jonathan~N Lee, George Tucker, Ofir Nachum, Bo~Dai, and Emma Brunskill.
\newblock Oracle inequalities for model selection in offline reinforcement learning.
\newblock \emph{Advances in Neural Information Processing Systems}, 35:\penalty0 28194--28207, 2022.

\bibitem[Liu et~al.(2025)Liu, Zhao, Agarwal, Liu, Huang, Amortila, and Jiang]{liu2025model}
Pai Liu, Lingfeng Zhao, Shivangi Agarwal, Jinghan Liu, Audrey Huang, Philip Amortila, and Nan Jiang.
\newblock Model selection for off-policy evaluation: New algorithms and experimental protocol.
\newblock \emph{arXiv preprint arXiv:2502.08021}, 2025.

\bibitem[Marinov and Zimmert(2021)]{marinov2021pareto}
Teodor~Vanislavov Marinov and Julian Zimmert.
\newblock The pareto frontier of model selection for general contextual bandits.
\newblock \emph{Advances in Neural Information Processing Systems}, 34:\penalty0 17956--17967, 2021.

\bibitem[Mnih et~al.(2015)Mnih, Kavukcuoglu, Silver, Rusu, Veness, Bellemare, Graves, Riedmiller, Fidjeland, Ostrovski, et~al.]{mnih2015human}
Volodymyr Mnih, Koray Kavukcuoglu, David Silver, Andrei~A Rusu, Joel Veness, Marc~G Bellemare, Alex Graves, Martin Riedmiller, Andreas~K Fidjeland, Georg Ostrovski, et~al.
\newblock Human-level control through deep reinforcement learning.
\newblock \emph{nature}, 518\penalty0 (7540):\penalty0 529--533, 2015.

\bibitem[Pacchiano et~al.(2020{\natexlab{a}})Pacchiano, Dann, Gentile, and Bartlett]{pacchiano2020regret}
Aldo Pacchiano, Christoph Dann, Claudio Gentile, and Peter Bartlett.
\newblock Regret bound balancing and elimination for model selection in bandits and rl.
\newblock \emph{arXiv preprint arXiv:2012.13045}, 2020{\natexlab{a}}.

\bibitem[Pacchiano et~al.(2020{\natexlab{b}})Pacchiano, Dann, Gentile, and Bartlett]{pacchiano2020regretboundbalancingelimination}
Aldo Pacchiano, Christoph Dann, Claudio Gentile, and Peter Bartlett.
\newblock Regret bound balancing and elimination for model selection in bandits and rl, 2020{\natexlab{b}}.
\newblock URL \url{https://arxiv.org/abs/2012.13045}.

\bibitem[Pacchiano et~al.(2020{\natexlab{c}})Pacchiano, Phan, Abbasi~Yadkori, Rao, Zimmert, Lattimore, and Szepesvari]{pacchiano2020model}
Aldo Pacchiano, My~Phan, Yasin Abbasi~Yadkori, Anup Rao, Julian Zimmert, Tor Lattimore, and Csaba Szepesvari.
\newblock Model selection in contextual stochastic bandit problems.
\newblock \emph{Advances in Neural Information Processing Systems}, 33:\penalty0 10328--10337, 2020{\natexlab{c}}.

\bibitem[Pacchiano~Camacho(2021)]{pacchiano2021model}
Aldo Pacchiano~Camacho.
\newblock \emph{Model Selection for Contextual Bandits and Reinforcement Learning}.
\newblock PhD thesis, UC Berkeley, 2021.

\bibitem[Parker-Holder et~al.(2022)Parker-Holder, Rajan, Song, Biedenkapp, Miao, Eimer, Zhang, Nguyen, Calandra, Faust, et~al.]{parker2022automated}
Jack Parker-Holder, Raghu Rajan, Xingyou Song, Andr{\'e} Biedenkapp, Yingjie Miao, Theresa Eimer, Baohe Zhang, Vu~Nguyen, Roberto Calandra, Aleksandra Faust, et~al.
\newblock Automated reinforcement learning (autorl): A survey and open problems.
\newblock \emph{Journal of Artificial Intelligence Research}, 74:\penalty0 517--568, 2022.

\bibitem[Schulman et~al.(2017)Schulman, Wolski, Dhariwal, Radford, and Klimov]{schulman2017proximal}
John Schulman, Filip Wolski, Prafulla Dhariwal, Alec Radford, and Oleg Klimov.
\newblock Proximal policy optimization algorithms.
\newblock \emph{arXiv preprint arXiv:1707.06347}, 2017.

\bibitem[Tassa et~al.(2012)Tassa, Erez, and Todorov]{6386025}
Yuval Tassa, Tom Erez, and Emanuel Todorov.
\newblock Synthesis and stabilization of complex behaviors through online trajectory optimization.
\newblock In \emph{2012 IEEE/RSJ International Conference on Intelligent Robots and Systems}, pages 4906--4913, 2012.
\newblock \doi{10.1109/IROS.2012.6386025}.

\bibitem[Watkins and Dayan(1992)]{watkins1992q}
Christopher~JCH Watkins and Peter Dayan.
\newblock Q-learning.
\newblock \emph{Machine learning}, 8\penalty0 (3):\penalty0 279--292, 1992.

\bibitem[Williams(1992)]{williams1992simple}
Ronald~J Williams.
\newblock Simple statistical gradient-following algorithms for connectionist reinforcement learning.
\newblock \emph{Machine learning}, 8\penalty0 (3):\penalty0 229--256, 1992.

\bibitem[Zhang et~al.(2024)Zhang, Hong, Pacchiano, and Agrawal]{zhang2024orso}
Chen Bo~Calvin Zhang, Zhang-Wei Hong, Aldo Pacchiano, and Pulkit Agrawal.
\newblock Orso: Accelerating reward design via online reward selection and policy optimization.
\newblock \emph{arXiv preprint arXiv:2410.13837}, 2024.

\end{thebibliography}
\bibliographystyle{plainnat}

\appendix
\section*{Appendix}
\startcontents[appendix]
\section*{Table of contents}
\printcontents[appendix]{}{1}{\setcounter{tocdepth}{2}}
\newpage
\section{Theoretical Results}
\label{sec::theory_appendix}

\subsection{Misspecification Test}
\label{sec::missp_test_exp}
The misspecification test determines whether the bound $\hat{d}_t^i \sqrt{n_t^i}$ matches the realized performance of the agent in a principled manner. For any base agents $j \in [M]$,
\begin{align*}
    v^* & \geq \frac{\bar{u}_t^j}{n_t^j} \tag{definition of $v^*$} \\
    & \geq \frac{u_t^j}{n_t^j} - c \sqrt{\ln \frac{\frac{M \ln n_t^j}{\delta}}{n_t^j}} \tag{Event $\mathcal{E}$}
\end{align*}

For a well-specified base agent $i \in [M]$ that satisfies its regret upper bound, 
\begin{align*}
    \Regret_t^i = n_t^i v^* - \bar{u}_t^i \leq \hat{d}_t^i \sqrt{n_t^i}
\end{align*}
Rearranging,
\begin{align*}
    v^* &\leq \frac{\bar{u}_t^i}{n_t^i} + \frac{\hat{d}_t^{i} \sqrt{n_t^{i}}}{n_t^{i}} \\
    &\leq \frac{u_t^i}{n_t^i} + c \sqrt{\ln \frac{\frac{M \ln n_t^j}{\delta}}{n_t^j}} + \frac{\hat{d}_t^{i} \sqrt{n_t^{i}}}{n_t^{i}} \tag{Event $\mathcal{E}$}
\end{align*}

Therefore, a well-specified agent $i$ should satisfy,
\begin{align*}
    \frac{u_t^i}{n_t^i} + c \sqrt{\ln \frac{\frac{M \ln n_t^i}{\delta}}{n_t^i}} + \frac{\hat{d}_t^{i} \sqrt{n_t^{i}}}{n_t^{i}}  \geq \frac{u_t^j}{n_t^j} - c \sqrt{\ln \frac{\frac{M \ln n_t^j}{\delta}}{n_t^j}} \quad\quad \forall j \in [M] 
\end{align*}
and otherwise the agent is misspecified, resulting in test \ref{eq::missp_test}. Triggering this misspecification test implies that the bound $\hat{d}_t^i \sqrt{n_t^i}$ is too small to bound the realized regret of the agent and hence he algorithm doubles the estimated regret coefficient $\hat{d}_t^i$. 

\subsection{Proof of Theorem \ref{theorem:: D3RB resource allocation}}
\label{sec::proof_D3RB_resource_allocation}
We reuse the following lemmas from \citep{dann2024data} to prove Theorem \ref{theorem:: D3RB resource allocation}.


\begin{lemma}
    \label{lemma::regret_multiplier}
    In event $\mathcal{E}$ \ref{eq::good_event}, for each base agent $i \in [M]$ , the regret multiplier $\hat{d}_t^i$ in algorithm \ref{alg::ModselRL}-left satisfies,
    \begin{equation}
        \hat{d}_t^i \leq 2 d_t^i, \quad\quad  \forall t \in \mathbb{N}
    \end{equation}
\end{lemma}

\begin{lemma}
    \label{lemma::balancing_potential}
    The potentials in Algorithm \ref{alg::ModselRL}-(left) are balanced at all times up to a factor 3, that is for all $t \in [T]$, 
    \begin{align}
    \phi_t^i \leq 3\phi_t^j \quad\quad \forall i, j \in [M] 
\end{align}
\end{lemma}

\textbf{ Proof of Theorem \ref{theorem:: D3RB resource allocation}} By Lemma \ref{lemma::balancing_potential}, the potentials in algorithm satisfy,
\begin{align}
    \phi_t^i \leq 3\phi_t^j \quad\quad \forall i, j \in [M] 
\end{align}
Substitute the definition of $\phi_t^i$,
\begin{align}
    \hat{d}_t^i \sqrt{n_t^i} & \leq 3 \hat{d}_t^j \sqrt{n_t^j} 
    \Rightarrow (\hat{d}_t^i)^2 n_t^i  \leq 9 (\hat{d}_t^j)^2 n_t^j
\end{align}
Rearrange,
\begin{align}
     n_t^i & \leq 9 \frac{(\hat{d}_t^j)^2}{ (\hat{d}_t^i)^2} n_t^j 
\end{align}
By lemma \ref{lemma::regret_multiplier}, we have $\hat{d}_t^i \leq 2d_t^i$,
\begin{align}
     & \leq  9 \frac{(d_t^j)^2}{(d_t^i)^2} n_t^j = 9 \frac{(1/d_t^i)^2}{(1/d_t^j)^2} n_t^j
\end{align}
For all $t\in[T]$ and a fixed $j \in [M]$,  
\begin{align}
    t & = \sum_{i=1}^M n_t^i 
    = n_t^j \left(\sum_{i=1}^{M}  \frac{(1/d_t^i)^2}{(1/d_t^j)^2}\right) 
    = \frac{n_t^j}{(1/d_t^j)^2} \sum_{i=1}^M (1/d_t^i)^2
\end{align}
Rearranging yields the final result in theorem \ref{theorem:: D3RB resource allocation},
\begin{align}
    n_t^j = \frac{(1/d_t^j)^2}{\sum_{i=1}^M (1/d_t^i)^2} \, t
\end{align}

\subsection{Proof of Theorem \ref{theorem::self-model-selection}}
\label{sec::proof_self_model_selection}
\textbf{ Proof of Theorem \ref{theorem::self-model-selection}}: Suppose agent $\mathcal{B}$ satisfies its theoretical upper bound with probability at least $1-\gamma(\delta)$, 

\begin{equation}      
\mathbb{P}\left(\Regret_t^\mathcal{B} \leq \mathcal{R}_\star(t, \delta) \quad \forall t \in [T]  \right) \geq 1 - \gamma(\delta)
\end{equation}

Then, if we combine $M$ independent base agents of type $\mathcal{B}$, the probability of at least one of them succeeding is larger than $1 - \gamma(\delta)^M$. Therefore, self-model selection requires M base agents to achieve its bound with probability $1 - \delta$,

\begin{equation}
    1 - \gamma(\delta)^M = 1 - \delta \Rightarrow M = \lceil\frac{\log\left(\delta\right)}{\log (\gamma(\delta))} \rceil
\end{equation}

\subsection{Sharing data via Importance sampling}
\label{sec::sharing_data}
\begin{theorem}
Consider the case of episodic RL, where we roll out policy $\pi^i$, and collect a trajectory $\tau = (s_1, a_1, r_1, \ldots, s_T,a_T, r_T) \sim \mathbb{P}^i(\tau) $. Here,  $\mathbb{P}^i(\tau)$ is the joint distribution over the trajectory $\tau$ under policy $\pi^i$,

\begin{align*}
    \mathbb{P}^i(\tau) = \mathcal{D}(s_1) \pi^i(a_1 | s_1) \mathcal{R}(r_1 | s_1, a_1) \mathcal{P}(s_2| s_1, a_1) \ldots \pi^i(a_T | s_T) \mathcal{R}(r_T | s_T, a_T)
\end{align*}

Then, $$\frac{\mathbb{P}^j(\tau)}{\mathbb{P}^i(\tau)} \sum_{t=1}^T \gamma^t r_t $$
is an unbiased estimate of the discounted episodic reward that we would've collected under $\pi^j$.

\proof 
By definition,
\begin{align*}
    \mathcal{J}(\pi^i) = \mathbb{E} \left[ \sum_{t=1}^T \gamma^{t-1} r_t \mid \tau \sim \mathbb{P}^i \right] = \mathbb{E}_{\tau \sim \mathbb{P}^i} \left[\sum_{t=1}^T \gamma^{t-1} r_t \right]
\end{align*}

By importance sampling,
\begin{align*}
    & \mathbb{E}_{\tau \sim \mathbb{P}^i} \left[\frac{\mathbb{P}^j(\tau)}{\mathbb{P}^i(\tau)} \sum_{t=1}^T \gamma^t r_t\right] \\
    & = \sum_{\tau} \left( \frac{\mathbb{P}^j(\tau)}{\mathbb{P}^i(\tau)} \sum_{t=1}^T \gamma^t r_t \right) \mathbb{P}^i(\tau )
    = \sum_{\tau} \left( \mathbb{P}^j(\tau) \sum_{t=1}^{T} \gamma^t r_t \right) \\
    & = \mathbb{E}_{\tau \sim \mathbb{P}^j} \left[ \sum_{t=1}^T \gamma^t r_t \right] = \mathbb{E}\left[ \sum_{t=1}^T \gamma^{t-1} r_t \mid \tau \sim \mathbb{P}^j \right] = \mathcal{J}(\pi^j)
\end{align*}

\end{theorem}

\begin{proposition} For a given trajectory $\tau =\{(s_t, a_t, r_t)\}_{t=1}^T$ collected by policy $\pi^i$,
\begin{align*}
    (\Pi_t \alpha_t) \sum_{t=1}^T \gamma^t r_t, \quad \text{where} \quad \alpha_t = \frac{\pi^j(a_t | s_t)}{\pi^i(a_t | s_t)}
\end{align*}
is the importance sampling ratio for the discounted episodic reward that we would have collected under policy $\pi^j$.

\proof
\begin{align*}
    &\mathbb{E}_{\tau \sim \mathbb{P}^i} \left[\frac{\mathbb{P}^j(\tau)}{\mathbb{P}^i(\tau)} \sum_{t=1}^T \gamma^t r_t\right] \\
    & = \mathbb{E}_{\tau \sim \mathbb{P}^i} \left[\frac{\mathcal{D}(s_1) \pi^j(a_1 | s_1) \mathcal{R}(r_1 | s_1, a_1) \mathcal{P}(s_2| s_1, a_1) \ldots \pi^j(a_T | s_T) \mathcal{R}(r_T | s_T, a_T) }{\mathcal{D}(s_1) \pi^i(a_1 | s_1) \mathcal{R}(r_1 | s_1, a_1) \mathcal{P}(s_2| s_1, a_1) \ldots \pi^i(a_T | s_T) \mathcal{R}(r_T | s_T, a_T)} \sum_{t=1}^T \gamma^t r_t\right] \\
    & = \mathbb{E}_{\tau \sim \mathbb{P}^i} \left[\frac{\pi^j(a_1 | s_1)  \ldots \pi^j(a_T | s_T) }{\pi^i(a_1 | s_1)  \ldots \pi^i(a_T | s_T)} \sum_{t=1}^T \gamma^t r_t\right] = \mathbb{E}_{\tau \sim \mathbb{P}^i} \left[ (\Pi_t \alpha_t) \sum_{t=1}^T \gamma^t r_t \right] 
\end{align*}

\end{proposition}

\subsection{Misspecification Test based on Realized versus Expected performance}
\label{subsec::classic_missp_test}

We consider the role of the misspecification test in the performance of model selection algorithms. We analyze and compare two of the algorithms, D$^3$RB and Classic Balancing, that select base agents by performing a misspecification test.  

Recall the set of base agents, $\mathcal{B} =  \{\mathcal{B}^1, \ldots, \mathcal{B}^M \}$ and suppose $\mathcal{B}^i$ has a high-probability upper bound $\mathcal{R}^i(t, \delta)$, 
\begin{align*}
    \{\mathcal{R}^1(t, \delta), \ldots, \mathcal{R}^M(t, \delta) \}    
\end{align*}

Suppose $\mathcal{B}^i$ is an instance of a sequential decision-making algorithm with high-probability regret guarantee $\mathcal{R}^i(t, \delta)$. 
\begin{equation}
    \mathbb{P} \left[ \Regret_t^i \leq \mathcal{R}^i(t, \delta) \right] \geq 1 - \delta
\end{equation}

Using these theoretical regret bounds, we can perform a similar misspecification test to \ref{eq::missp_test},
\begin{equation}
    \label{eq::theoretical_missp_test}
    \frac{u_t^i}{n_t^i}  + c \sqrt{\ln \frac{\frac{M \ln n_t^i}{\delta}}{n_t^i}} + \frac{\mathcal{R}^i(t, \delta)}{n_t^i} \leq \max_{j \in [M]} \frac{u_t^j}{n_t^j} - c \sqrt{\ln \frac{\frac{M \ln n_t^j}{\delta}}{n_t^j}}
\end{equation}

At round $t$, the Classic Balancing algorithm selects the base agent with minimum regret bound $i_t = \arg\min_{i \in [M]} R(t, \delta)$, eliminating the base agents that are flagged as misspecified by test \ref{eq::theoretical_missp_test}. Note that, the difference between tests \ref{eq::missp_test} and \ref{eq::theoretical_missp_test} is that one is performed based on the theoretical bound, and the other is performed using the data-adaptive regret bound actively estimated by realized rewards.  The drastic difference between the performance of D$^3$RB and Classic Balancing selectors in figure \ref{fig::step_size_selection_baselines} highlights the role of designing data-adaptive model selection algorithms versus selectors that perform upon theoretical bounds on the expected performance of the agent. 
\newpage

\section{Experimental Details}
\label{sec::experiment_appendix}

\subsection{Step Size Selection}
\label{sec::step_size_selection_plots}
Here, we include the selection statistics plot for the rest of algorithms in experiment \ref{sec::step_size_selection_plots}.



\begin{figure}[htbp]
    \centering
    \includegraphics[width=0.3\textwidth]{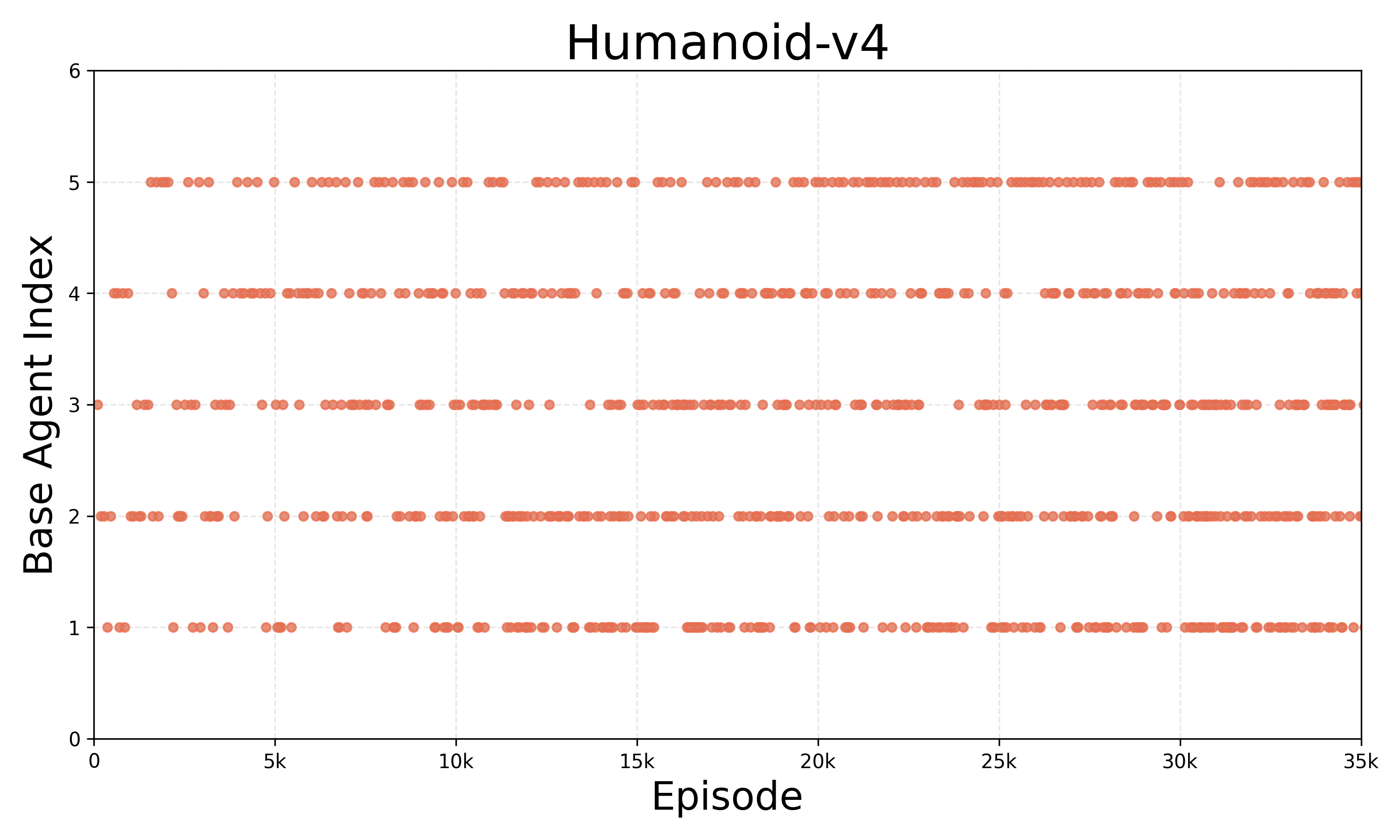}
    \hfill
    \includegraphics[width=0.3\textwidth]{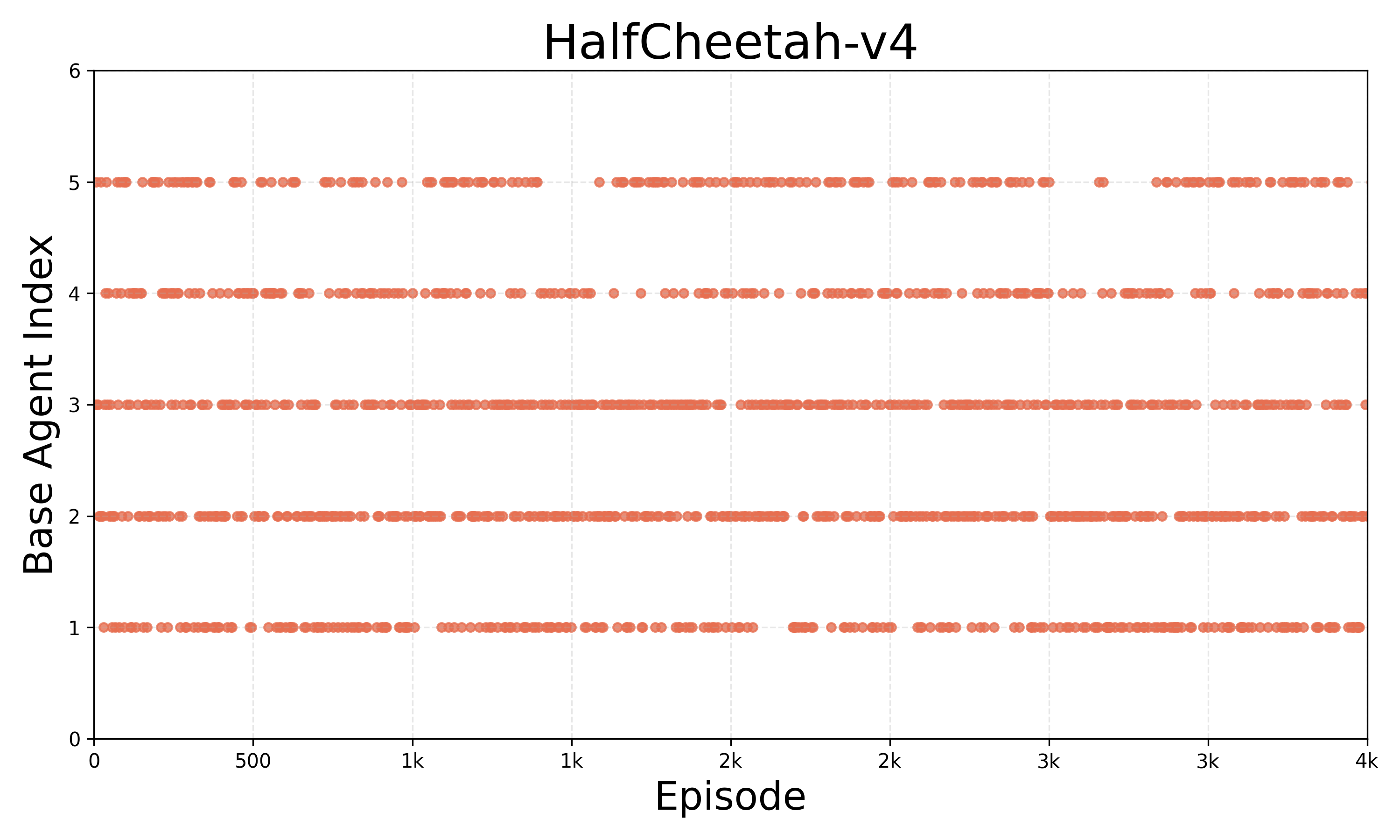}
    \hfill
    \includegraphics[width=0.3\textwidth]{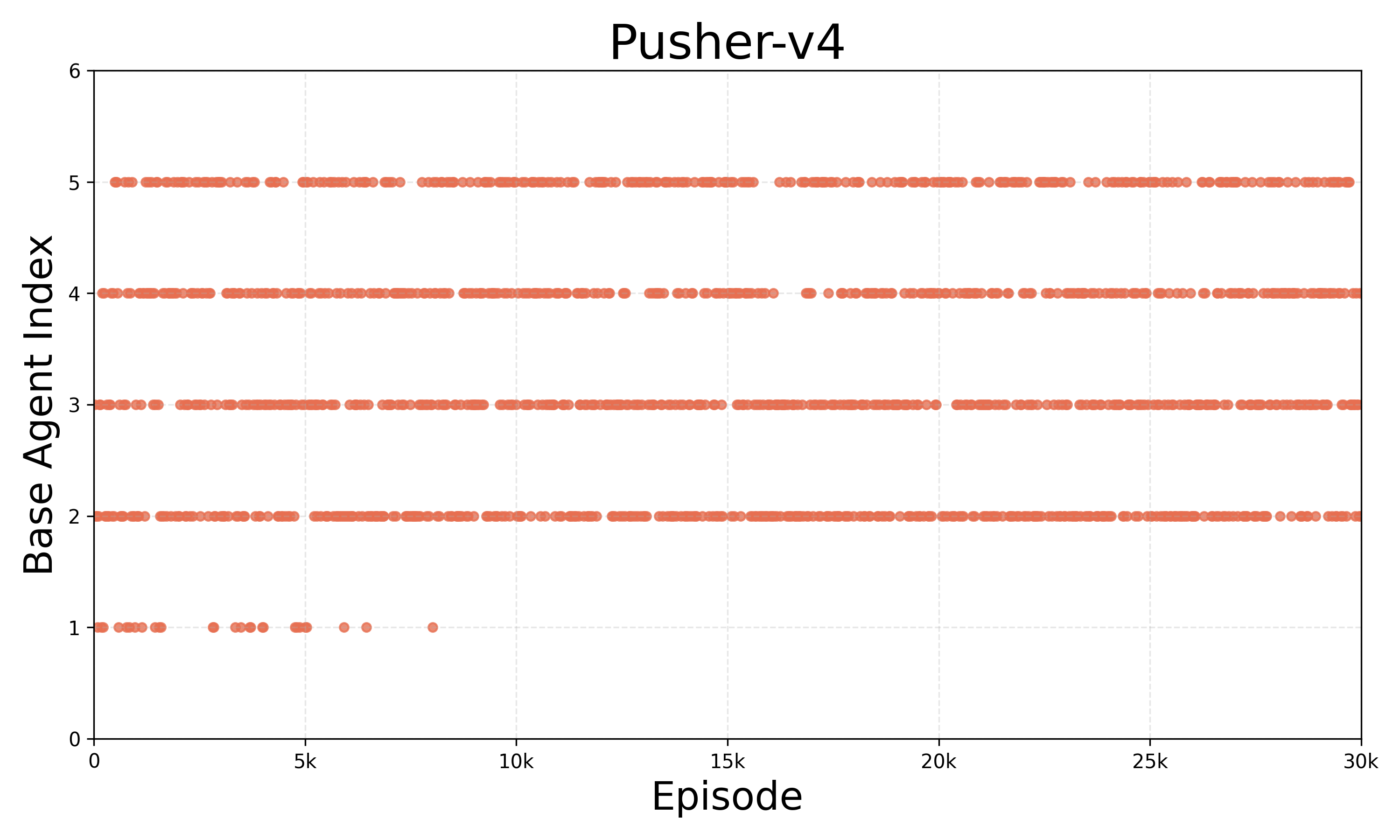}
    \caption{Exp3 Selection Statistics}
    \label{fig:three-figures}
\end{figure}

\begin{figure}[htbp]
    \centering
    \includegraphics[width=0.3\textwidth]{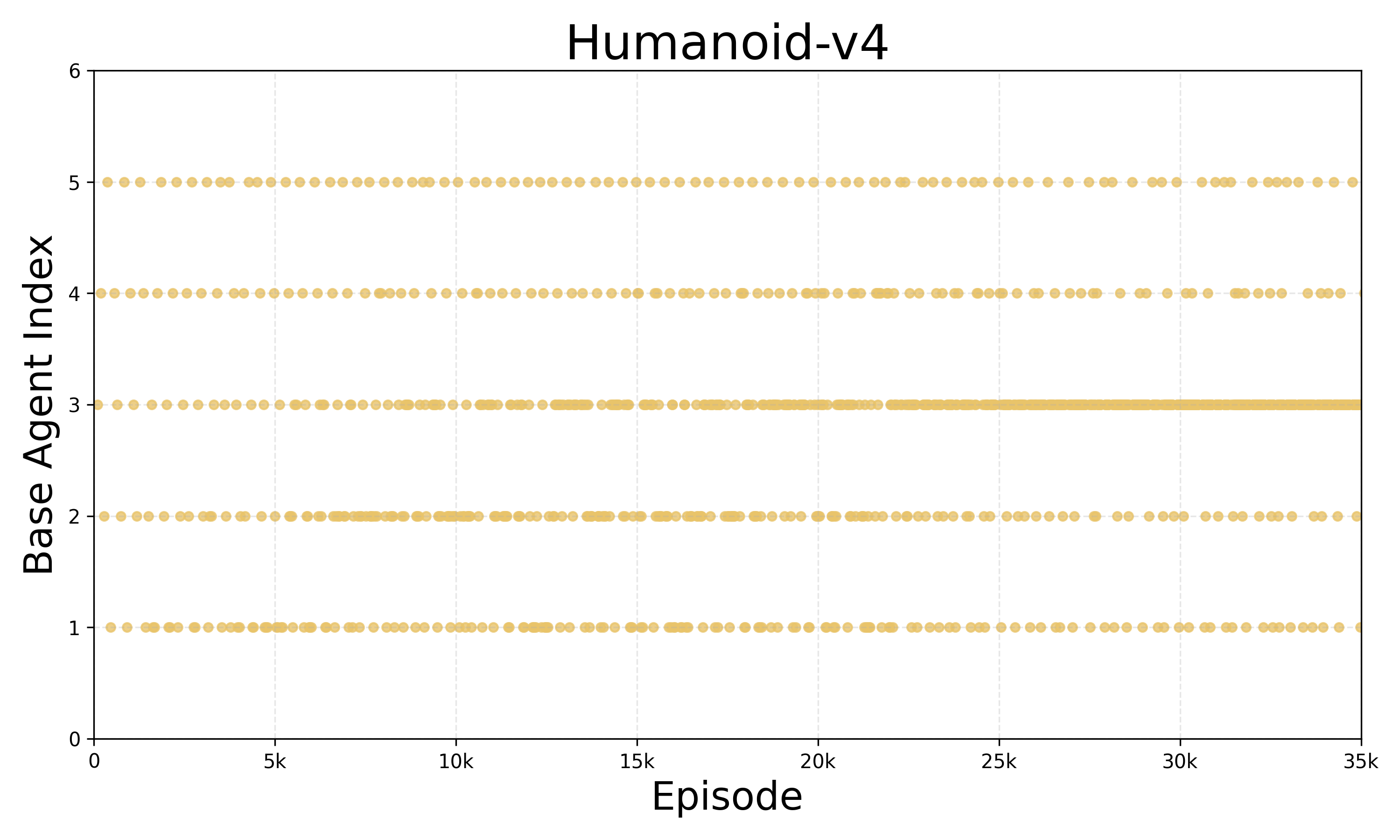}
    \hfill
    \includegraphics[width=0.3\textwidth]{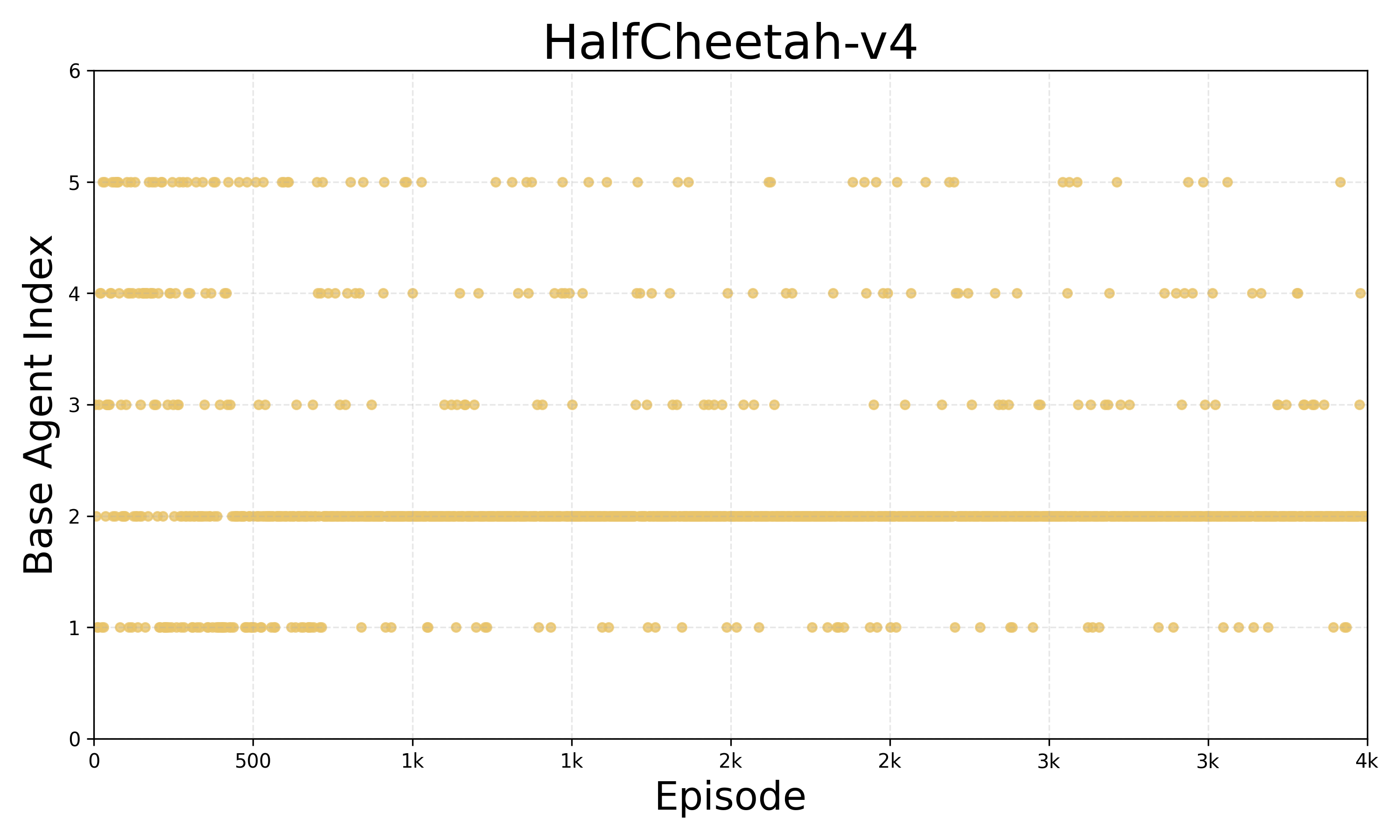}
    \hfill
    \includegraphics[width=0.3\textwidth]{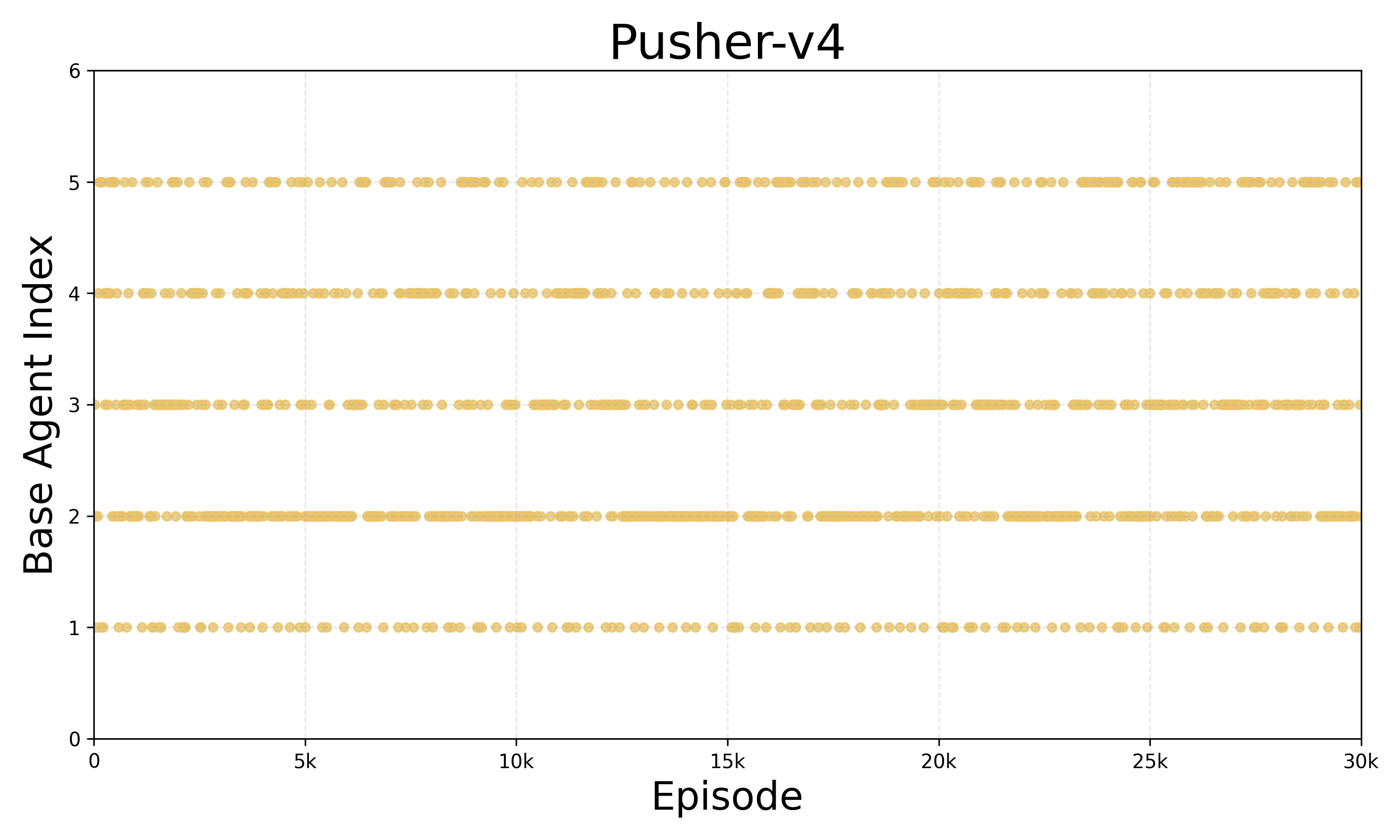}
    \caption{Corral Selection Statistics}
    \label{fig:three-figures}
\end{figure}
\begin{figure}[htbp]
    \centering
    \includegraphics[width=0.3\textwidth]{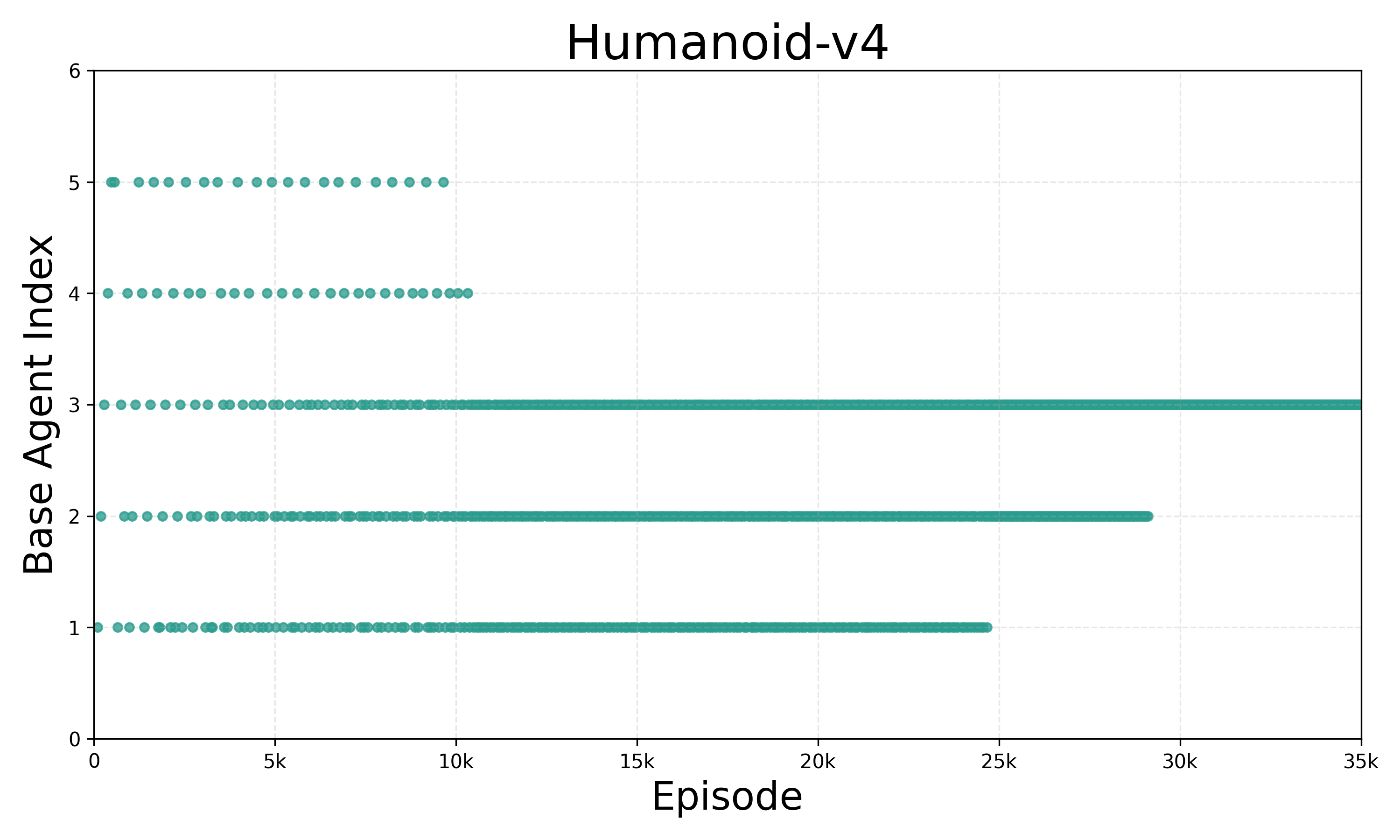}
    \hfill
    \includegraphics[width=0.3\textwidth]{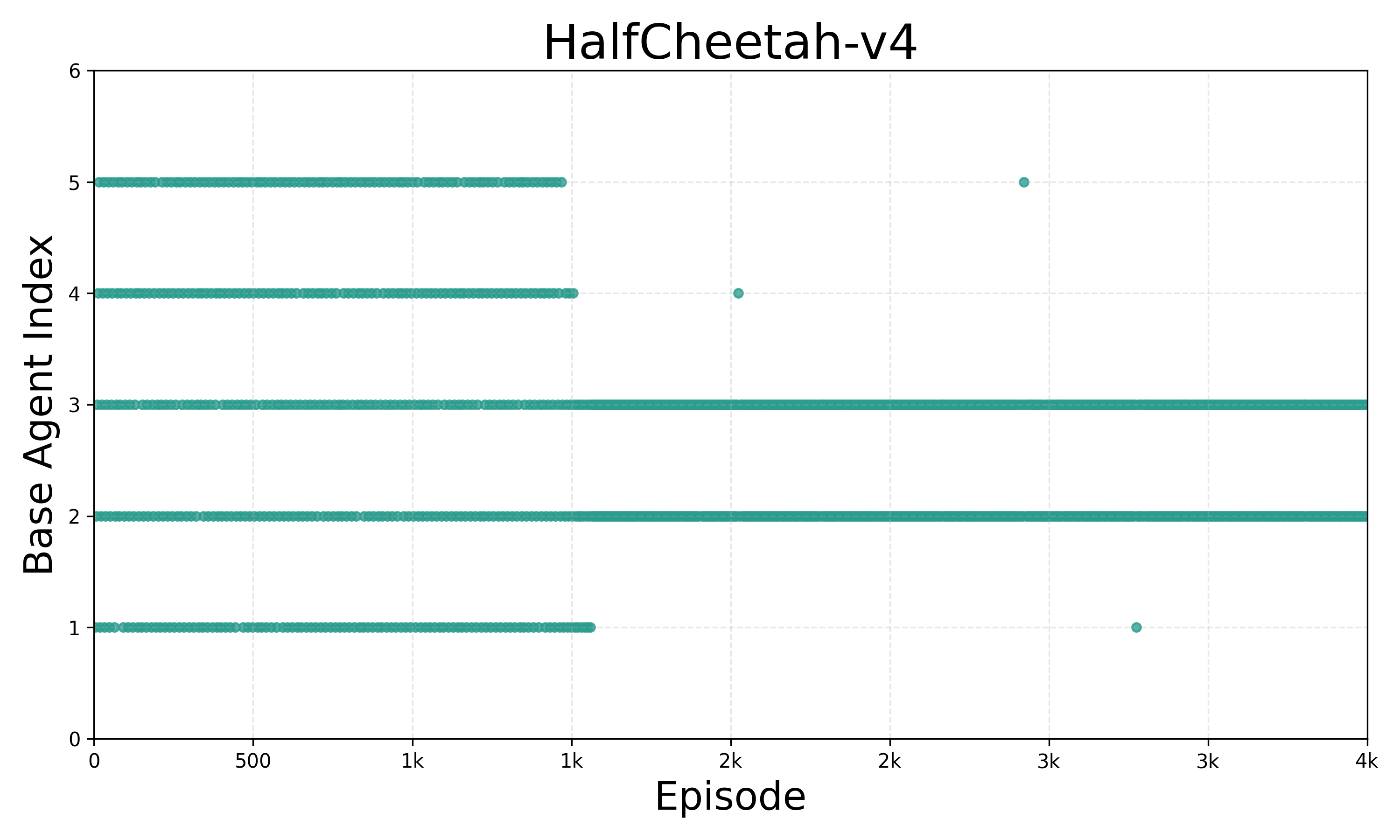}
    \hfill
    \includegraphics[width=0.3\textwidth]{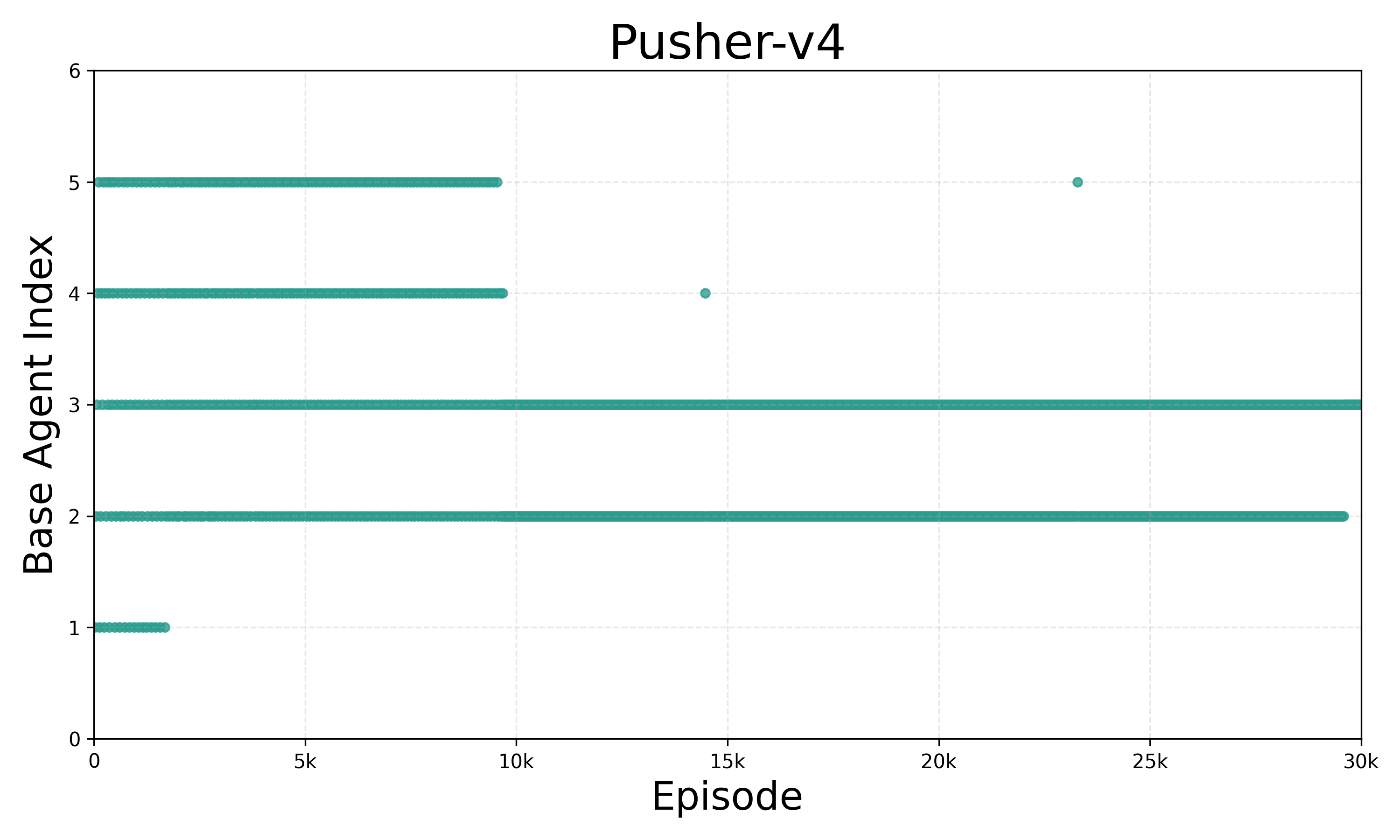}
    \caption{ED$^2$RB Selection Statistics}
    \label{fig:three-figures}
\end{figure}
\begin{figure}[htbp!]
    \centering
    \includegraphics[width=0.3\textwidth]{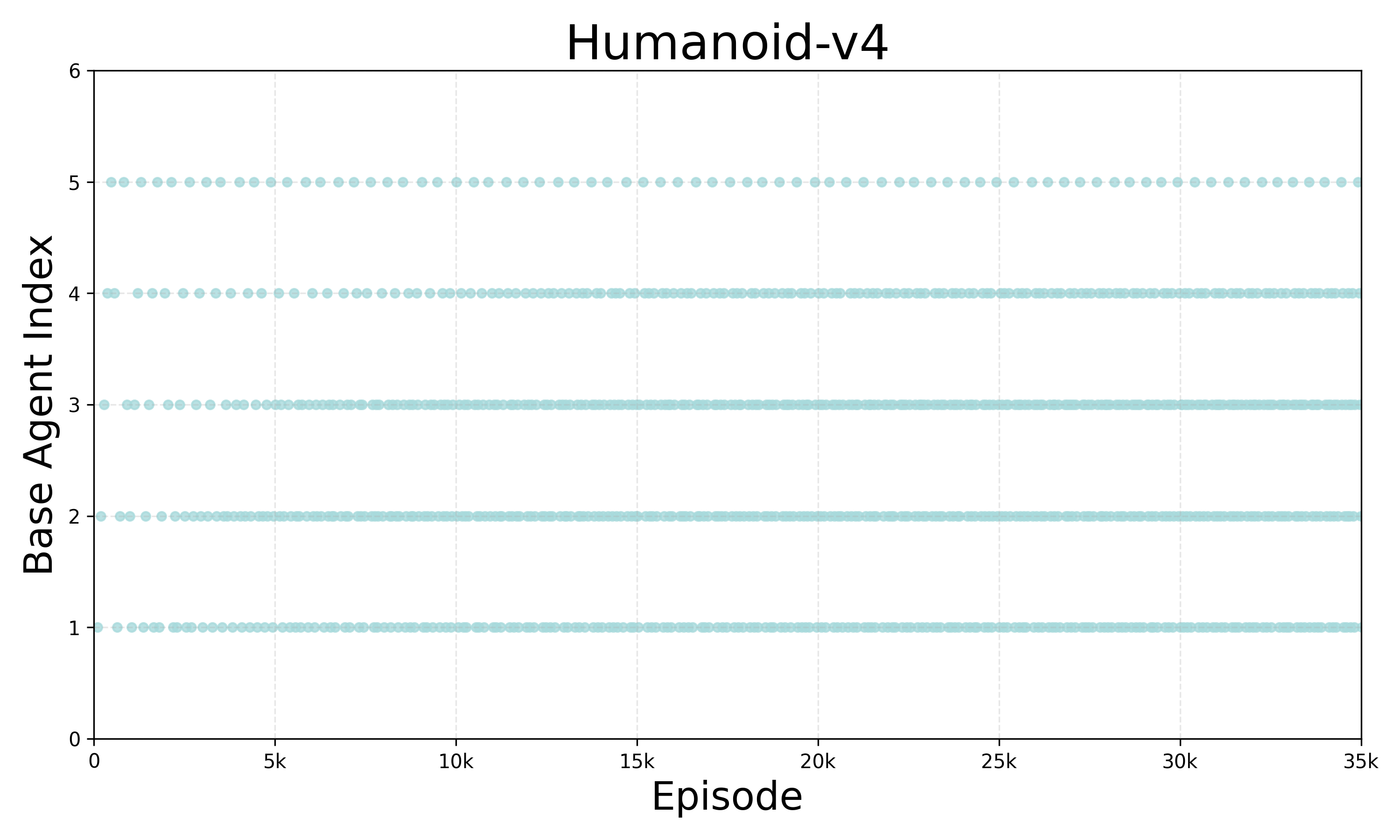}
    \hfill
    \includegraphics[width=0.3\textwidth]{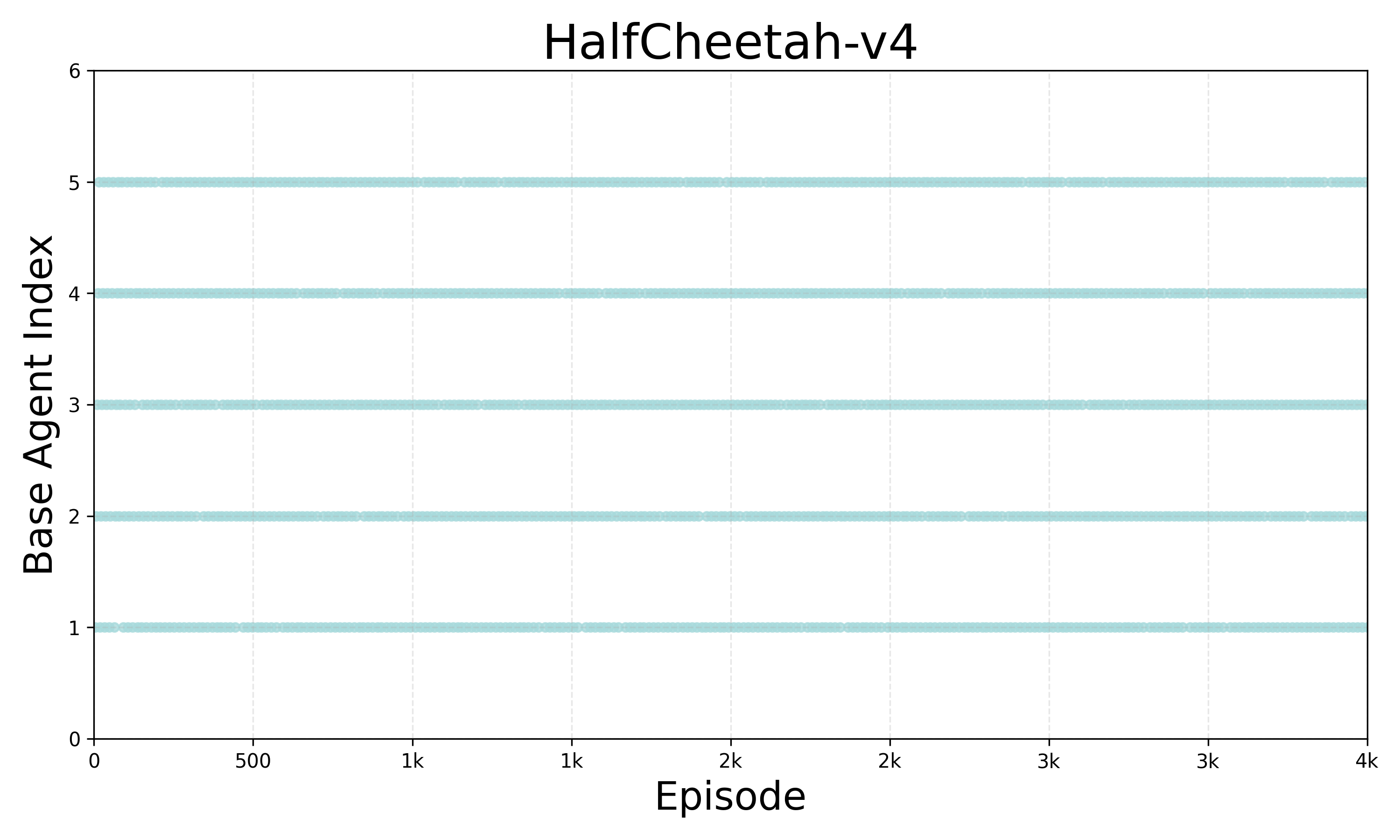}
    \hfill
    \includegraphics[width=0.3\textwidth]{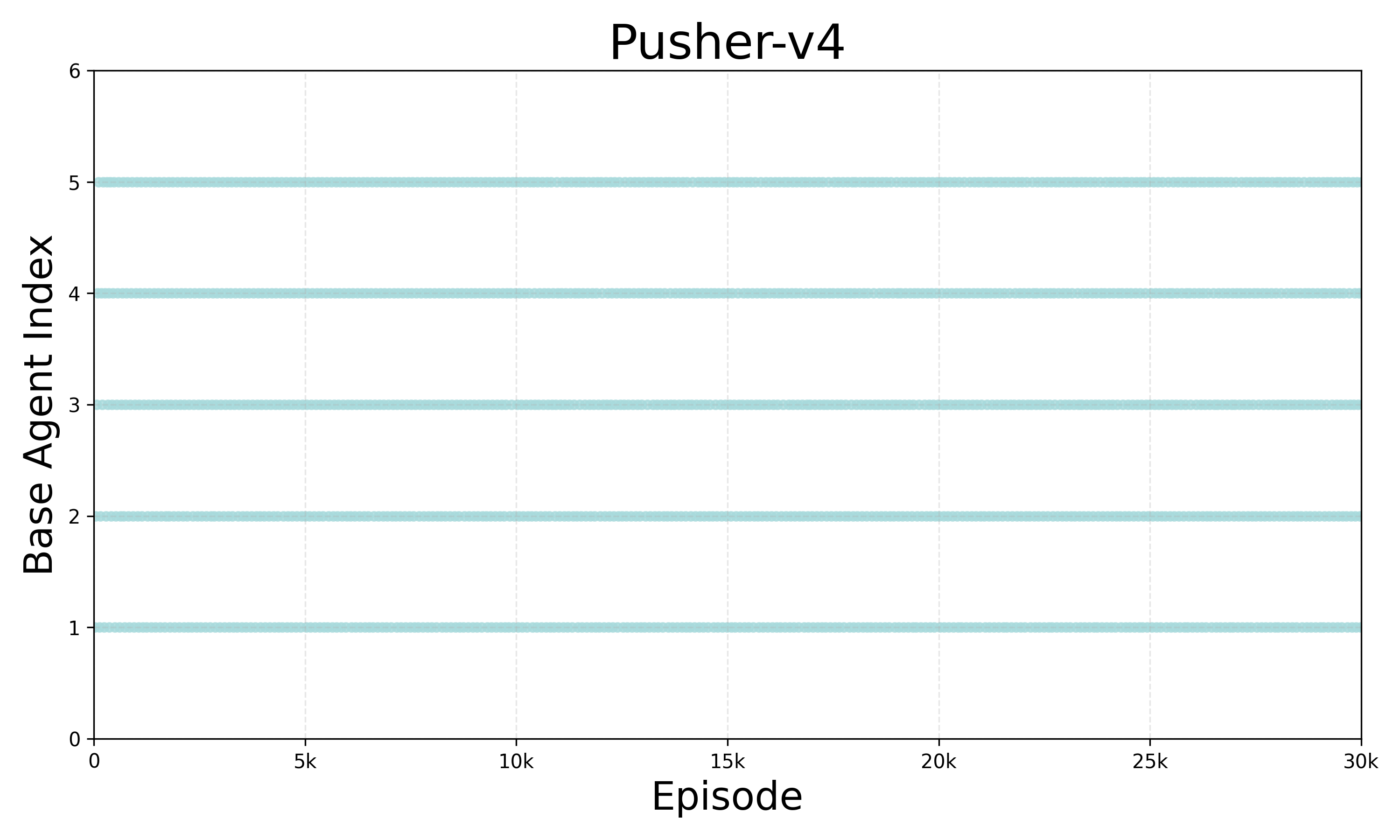}
    \caption{Classic Selection Statistics}
    \label{fig::classic_selection}
\end{figure}

\newpage

\subsection{Neural Architecture Selection}
\label{sec::neural_architectures_details}

The following is the detailed architectural choice of three base agents in experiment. \ref{sec::neural_architecture_selection}. The second architecture has the configuration with best realized performance. The first architecture, roughly shares the same number of parameters, but only has one convolutional layer, limiting the agents ability in tasks that require temporal reasoning. The third architecture shares that same number of layers to the second one, but has less parameters that limits the representational capacity of the agent.

$\mathcal{B}^1$ network architecture: 
\begin{lstlisting}[language=Python]
    network = nn.Sequential(       
                nn.Conv2d(4, 16, 8, stride=4),
                nn.ReLU(),
                nn.Flatten(),
                nn.Linear(6400,256),
                nn.ReLU(),
                nn.Linear(256, env.single_action_space.n),
            )
\end{lstlisting}

$\mathcal{B}^2$ network architecture: 
\begin{lstlisting}[language=Python]
     network = nn.Sequential(
                    nn.Conv2d(4, 32, 8, stride=4),
                    nn.ReLU(),
                    nn.Conv2d(32, 64, 4, stride=2),
                    nn.ReLU(),
                    nn.Conv2d(64, 64, 3, stride=1),
                    nn.ReLU(),
                    nn.Flatten(),
                    nn.Linear(3136, 512),
                    nn.ReLU(),
                    nn.Linear(512, env.single_action_space.n),
                )
\end{lstlisting}

$\mathcal{B}^3$ network architecture: 
\begin{lstlisting}[language=Python]
      network = nn.Sequential(        
                nn.Conv2d(4, 8, 8, stride=4),
                nn.ReLU(),
                nn.Conv2d(8, 8, 4, stride=2),
                nn.ReLU(),
                nn.Conv2d(8, 8, 3, stride=1),
                nn.ReLU(),
                nn.Flatten(),
                nn.Linear(392, 32),
                nn.ReLU(),
                nn.Linear(32, env.single_action_space.n),
            )
\end{lstlisting}

\newpage
\section{Algorithmic Details}

\subsection{Model Selection Algorithms}
\label{sec::appendix_modsel_alg}

In this section, we provide 5 model selection strategies that follow the interface in algorithm \ref{alg::ModselRL}-left. To avoid including all the theoretical details, there might be a slight abuse of notation in the pseudocodes.

\subsubsection*{ED$^2$RB}
 Estimating Data Driven Regret Balancing (ED$^2$RB) \citep{dann2024data} is similar to D$3$RB, though it tries to directly estimate the regret coefficients.

\begin{algorithm}[H]
\SetKwInput{KwInput}{Input}                
\SetKwInput{KwOutput}{Output}              
\DontPrintSemicolon
  
  \KwInput{$m$, $\beta$, $\Psi$, $\delta$}

  \SetKwFunction{FMain}{Main}
  \SetKwFunction{FSum}{sample}
  \SetKwFunction{FSub}{update}
 
  \SetKwProg{Fn}{Function}{:}{}
  \Fn{\FSum{}}{
        \tcp{Sample base index}
        $i = \argmin_j \Psi_j$\;
        $\pi_i, \alpha_i \leftarrow \beta_i$ \;
        \KwRet $i, \pi_i, \alpha_i$\;
  }

  \SetKwProg{Fn}{Function}{:}{}
  \Fn{\FSub{$i$, $R[1:T]$}}{ 
      $R_{norm} \leftarrow normalize (R[1:T])$ \;
      \tcp{Update Statistics}
        $u^i = u^i+R_{norm}$ \;
        $n^i = n^i + 1$ \;
    \tcp{Estimate active regret coefficient}
        $d^i = \max\{d_{min}, \sqrt{n_t^{i_t}}(max_j \frac{u^j}{n^j} - c \sqrt{ln \frac{\frac{M ln n^j}{\delta}}{n^j}} - c \sqrt{ln \frac{\frac{m ln n^i}{\delta}}{n^i}} - \frac{u^i}{n^i})\}
        $ \;
    \tcp{Update balancing potential}
        $\Psi^i = clip(d^i\sqrt{n^i}, \Psi^i, 2\Psi^i)$ \;
    
  }
\label{alg3}
\caption{ED$^2$RB}
\end{algorithm}

\subsubsection*{Classic Balancing}
The Classic Regret Balancing Algorithm \citep{pacchiano2020regret} starts with the full set of base agents $\beta = [\beta_1, ..., \beta_m]$, at each round, the algorithm performs miss-specification on each of the base agents and eliminates the miss-specified one. Denote $\Psi^j$ as the empirical regret upper bound of base agent $j$.

\begin{algorithm}[H]
\SetKwInput{KwInput}{Input}                
\SetKwInput{KwOutput}{Output}              
\DontPrintSemicolon
  
  \KwInput{$m$, $\beta$, $\Psi$, $\delta$}

  \SetKwFunction{FMain}{Main}
  \SetKwFunction{FSum}{sample}
  \SetKwFunction{FSub}{update}
 
  \SetKwProg{Fn}{Function}{:}{}
  \Fn{\FSum{}}{
        \tcp{Sample Base index}
        $i = \argmin_j \Psi_j$\;
        $\pi_i, \alpha_i \leftarrow \beta_i$ \;
        \KwRet $i, \pi_i, \alpha_i$\;
  }

  \SetKwProg{Fn}{Function}{:}{}
  \Fn{\FSub{$i$, $R[1:T]$}}{
      $R_{norm} \leftarrow normalize (R[1:T])$ \;
      
        \tcp{Update statistics}
        $u^i = u^i+R_{norm}$ \;
        $n^i = n^i + 1$ \;

        \tcp{Perform miss-specification test for all the remaining base agents}
        \For{$\beta_k\in \beta$}
        {$ \frac{u^k}{n^k} + \frac{d^k \sqrt{n^k}}{n^i} + c \sqrt{ln \frac{\frac{m ln n^k}{\delta}}{n^k}} \leq max_j \frac{u^j}{n^j} - c \sqrt{ln \frac{\frac{M ln n^j}{\delta}}{n^j}}$ \;
        \If{miss-specified}{
            $\beta \leftarrow \beta / \{\beta_{k}\}$ \;
        } \;
        }

  }
\label{alg4}
\caption{Classic Balancing}
\end{algorithm}

\subsubsection*{EXP3}
Exponential-weight algorithm for exploration and exploitation (EXP3) learns a probability distribution $\Psi^i = \frac{exp(S^i)}{\sum_{j=1}^{m} exp(S^j)}$ over base learners, where $S^i$ is a total estimated reward of base agent $i$ up to this round.

\begin{algorithm}[H]
\SetKwInput{KwInput}{Input}                
\SetKwInput{KwOutput}{Output}              
\DontPrintSemicolon
  
  \KwInput{$m$, $\beta$, $\Psi$, $\delta$}

  \SetKwFunction{FMain}{Main}
  \SetKwFunction{FSum}{sample}
  \SetKwFunction{FSub}{update}
 
  \;
  \SetKwProg{Fn}{Function}{:}{}
  \Fn{\FSum{}}{
        \tcp{Sample Base index}
        $i = \argmax_j \Psi_j$\;
        
        $\pi_i, \alpha_i \leftarrow \beta_i$ \;
        \;
        \KwRet $i, \pi_i, \alpha_i$\;
  }
  \SetKwProg{Fn}{Function}{:}{}
  \Fn{\FSub{$i$, $R[1:T]$}}{ \;
      $R_{norm} \leftarrow normalize (R[1:T])$ \;
      \tcp{Update statistics}
        
        \For{$j\in {1, ..., m}$} 
          {$S^j = S^j+ 1 - \frac{\mathbb{I}\{j = i\}(1-R_{norm})}{\Psi^i}$}
        \;
    \tcp{Update Distribution}
        $\Psi^i = \frac{exp(S^i)}{\sum_{j=1}^{m} exp(S^j)}$ \;
    
  }
\label{alg5}
\caption{EXP3}
\end{algorithm}

\subsubsection*{Corral}
Corral \citep{agarwal2017corralling} learns a distribution $\Psi$ over base agents and updates it according to LOG-BARRIER-OMD algorithm. We skip the algorithmic details and refer to the updating rule mentioned in the original paper as Corral-Update.

\begin{algorithm}[H]
\SetKwInput{KwInput}{Input}                
\SetKwInput{KwOutput}{Output}              
\DontPrintSemicolon
  
  \KwInput{$m$, $\beta$, $\Psi$}

  \SetKwFunction{FMain}{Main}
  \SetKwFunction{FSum}{sample}
  \SetKwFunction{FSub}{update}
 
  \;
  \SetKwProg{Fn}{Function}{:}{}
  \Fn{\FSum{}}{
        \tcp{Sample base index}
        $i \sim \Psi$\;
        
        $\pi_i, \alpha_i \leftarrow \beta_i$ \;
        \;
        \KwRet $i, \pi_i, \alpha_i$\;
  }

  \SetKwProg{Fn}{Function}{:}{}
  \Fn{\FSub{$i$, $R[1:T]$}}{ \;
      $R_{norm} \leftarrow normalize (R[1:T])$ \;
      \tcp{Update according to Corral}
        $\Psi^j \leftarrow \text{Corral-Update}(R_{norm})$ \;
    
  }
\label{alg6}
\caption{Corral}
\end{algorithm}

\subsubsection*{UCB}
The Upper Confidence Bound algorithm (UCB) maintains an optimistic estimate of the mean for each arm \citep{lattimore2020bandit}. Denote $\Psi^i$ as the upper confidence bound of arm $i$. The UCB algorithm  for learning rate-free RL works as follows,

\begin{algorithm}[H]
\SetKwInput{KwInput}{Input}                
\SetKwInput{KwOutput}{Output}              
\DontPrintSemicolon
  
  \KwInput{$m$, $\beta$, $\Psi$, $\delta$}

  \SetKwFunction{FMain}{Main}
  \SetKwFunction{FSum}{sample}
  \SetKwFunction{FSub}{update}
 
  \SetKwProg{Fn}{Function}{:}{}
  \Fn{\FSum{}}{
        \;
        \tcp{Sample base index}
        $i = \argmax_j \Psi_j$\;
        
        $\pi_i, \alpha_i \leftarrow \beta_i$ \;
        \;
        \KwRet $i, \pi_i, \alpha_i$\;
  }
  
  \SetKwProg{Fn}{Function}{:}{}
  \Fn{\FSub{$i$, $R[1:T]$}}{ \;
      $R_{norm} \leftarrow normalize (R[1:T])$ \;
      \tcp{Update statistics}
        $u^i = u^i+R_{norm}$ \;
        $n^i = n^i + 1$ \;
        $\mu^i = \frac{u^i}{n^i}$
        
    \tcp{Update Upper Confidence Bounds}
        $\Psi^i = UCB^i (\delta) =  \mu^i + \sqrt{\frac{2 log(1 / \delta)}{n^i}}$ \;
    
  }
\label{alg7}
\caption{UCB}
\end{algorithm}

\subsection{RL Training Updates}
\label{sec::appendix_rl_alg}
Two of the predominant approaches for learning the (near) optimal policy in reinforcement learning are policy optimization and Q-learning. Policy optimization starts with an initial policy and in each episode updates the parameters 
by taking gradient steps toward maximizing the episodic return. Denote learning rate as $\alpha \in \mathbb{R}$, a common update rule in policy optimization methods is
\begin{equation}
    \theta \leftarrow \theta + \alpha \: \mathbb{E} \biggl[ \sum_{t=0}^{T} \nabla_{\theta} \log \pi_{\theta} (s_t, a_t) (Q^{\pi_{\theta}}(s_t, a_t) - V^{\pi_{\theta}}(s_t)) \biggr]
\end{equation}

Q-learning uses the temporal differences method to update the parameters of $Q^{\pi_{\theta}}$. A common update rule is
\begin{equation}
    \theta \leftarrow \theta + \alpha  \: \mathbb{E}_{s,a,s^{\prime}, r \sim D} \biggl[\nabla_{\theta} (r + \gamma \: \max_{a^{\prime} \in A} Q^{\pi_{\bar\theta}}(s^{\prime}, a^{\prime}) - Q^{\pi_{\theta}}(s, a))^2 \biggr]
\end{equation}
where $D$ is the experience replay buffer and $\bar{\theta}$ is a frozen parameter set named target parameter. Proximal Policy Optimization (PPO) \citep{schulman2017proximal} and Deep Q-Networks (DQN) \citep{mnih2015human} follow the first and second approaches, respectively.

\end{document}